\documentclass{article}

 \usepackage[main, final, nonatbib]{neurips_2026}

\usepackage[utf8]{inputenc} %
\usepackage[T1]{fontenc}    %
\usepackage{url}            %
\usepackage{booktabs}       %
\usepackage{amsfonts}       %
\usepackage{nicefrac}       %
\usepackage{microtype}      %
\usepackage{xcolor}         %
\usepackage{paralist}

\usepackage[numbers]{natbib} %
\usepackage[small]{caption}
\usepackage{booktabs}
\usepackage{multirow}
\usepackage{longtable}
\usepackage{soul}
\usepackage[table]{xcolor}
\usepackage{enumitem}
\usepackage{tcolorbox}
\usepackage{wrapfig}
\usepackage{subcaption}
\usepackage{siunitx}
\usepackage[colorlinks=true,allcolors=blue,final]{hyperref}
\usepackage{xspace}
\usepackage{placeins}

\usepackage{tikz}
\usetikzlibrary{matrix,positioning,arrows.meta,fit,tikzmark,backgrounds}

\usepackage{amsmath,bm}
\usepackage{amssymb}
\usepackage{mathtools}
\usepackage{amsthm}
\usepackage{thmtools}

\def\Figref#1{Figure~\ref{#1}}

\def\Secref#1{Section~\ref{#1}}

\def\1{\bm{1}}

\def\vzero{{\bm{0}}}

\def\vmu{{\bm{\mu}}}
\def\vtheta{{\bm{\theta}}}

\def\vb{{\bm{b}}}
\def\vc{{\bm{c}}}
\def\vd{{\bm{d}}}

\def\vu{{\bm{u}}}
\def\vv{{\bm{v}}}
\def\vw{{\bm{w}}}
\def\vx{{\bm{x}}}

\def\vz{{\bm{z}}}

\def\mA{{\bm{A}}}

\def\mC{{\bm{C}}}

\def\mE{{\bm{E}}}

\def\mI{{\bm{I}}}

\def\mL{{\bm{L}}}

\def\mO{{\bm{O}}}

\def\mR{{\bm{R}}}

\def\mW{{\bm{W}}}

\def\mPhi{{\bm{\Phi}}}

\def\mSigma{{\bm{\Sigma}}}

\DeclareMathAlphabet{\mathsfit}{\encodingdefault}{\sfdefault}{m}{sl}
\SetMathAlphabet{\mathsfit}{bold}{\encodingdefault}{\sfdefault}{bx}{n}

\def\gD{{\mathcal{D}}}

\def\gL{{\mathcal{L}}}

\def\gN{{\mathcal{N}}}
\def\gO{{\mathcal{O}}}

\def\gQ{{\mathcal{Q}}}

\newcommand{\E}{\mathbb{E}}

\newcommand{\R}{\mathbb{R}}

\newcommand{\KL}{D_{\mathrm{KL}}}

\DeclareMathOperator*{\argmin}{arg\,min}

\newtheorem{corollary}{Corollary}

\newtheorem{proposition}{Proposition}
\newtheorem*{proposition*}{Proposition}
\newtheorem{remark}{Remark}

\newcommand{\defeq}{\vcentcolon=}

\newcommand{\viposterior}{\ensuremath{q(\mW \mid \bm{\eta})}}

\newcommand{\rev}[1]{{\color{magenta}#1}}
\newcommand{\selbo}{\texttt{s-NLL}\xspace}
\newcommand{\melbo}{\texttt{m-NLL}\xspace}
\newcommand{\mfcrps}{\texttt{m-CRPS}\xspace}

\newcommand{\marcrps}{\texttt{m-arCRPS}\xspace}
\newcommand{\marnll}{\texttt{m-arNLL}\xspace}
\newcommand{\mkde}{\texttt{m-KDE}\xspace}

\makeatletter
\DeclareRobustCommand\onedot{\futurelet\@let@token\@onedot}
\def\@onedot{\ifx\@let@token.\else.\null\fi\xspace}

\def\eg{\emph{e.g}\onedot} 
\def\ie{\emph{i.e}\onedot} 
\def\cf{\emph{cf}\onedot}

\makeatother

\newcommand{\fakepar}[1]{\vspace{1mm}\noindent\textbf{#1}}
\renewcommand{\paragraph}[1]{\par\fakepar{#1}\ }

\newtcbox{\methodbox}[1][]{%
  on line,
  arc=1.5pt,
  boxrule=0pt,
  left=0.1pt,
  right=0.1pt,
  top=0.1pt,
  bottom=0.1pt,
  colback=#1,
}

\definecolor{tabblue}{RGB}{31,119,180}
\definecolor{taborange}{RGB}{255,127,14}
\definecolor{tabgreen}{RGB}{44,160,44}
\definecolor{tabred}{RGB}{214,39,40}
\definecolor{tabpurple}{RGB}{148,103,189}
\definecolor{tabbrown}{RGB}{140,86,75}
\definecolor{tabpink}{RGB}{227,119,194}
\definecolor{tabgray}{RGB}{127,127,127}
\definecolor{tabolive}{RGB}{188,189,34}
\definecolor{tabcyan}{RGB}{23,190,207}

\newcommand{\cselbo}{\methodbox[tabred!20]{\selbo}}
\newcommand{\cmkde}{\methodbox[tabgreen!20]{\mkde}}
\newcommand{\cmelbo}{\methodbox[tabblue!20]{\melbo}}
\newcommand{\cmfcrps}{\methodbox[taborange!20]{\mfcrps}}
\newcommand{\cmarcrps}{\methodbox[tabpurple!20]{\marcrps}}
\newcommand{\cmarnll}{\methodbox[tabgray!20]{\marnll}}

\title{Gibbs Variational Last Layers for Long-Horizon Dynamics Learning}

\author{%
  Paul Brunzema \quad Henrik Hose \quad Sebastian Trimpe \\
  RWTH Aachen University\\
  Aachen, Germany \\
  \texttt{\{firstname.lastname\}@dsme.rwth-aachen.de} \\
}

\begin{document}
\maketitle

\begin{abstract}

    Accurate long-horizon probabilistic predictions of dynamical systems are a prerequisite for model-based decision-making.
    Variational Bayesian last layers (VBLLs) are an attractive model class for this setting, offering tractable uncertainty at near-deterministic cost, yet they are trained exclusively with single-step likelihood objectives that may be insufficient for reliable multi-step rollouts.
    We revisit VBLLs through the lens of Gibbs variational inference, which decouples posterior inference from the choice of training loss%
    .
    Building on this perspective, we introduce a family of multi-step training losses that progressively incorporate long-horizon structure, robustness via continuous ranked probability scores (CRPS), and training on full autoregressive rollouts.
    Our approach retains the simplicity
    of VBLLs while directly optimizing for multi-step predictive quality.
    Across illustrative synthetic environments, a diverse suite of chaotic dynamical systems, and real-world data, we show that multi-step, CRPS-based objectives substantially improve long-horizon accuracy and distributional fit, consistently outperforming single-step likelihood training.
\end{abstract}

\section{Introduction}
\label{sec:intro}
Learned probabilistic dynamics models underpin a growing range of decision-making systems, from sampling-based control \citep{levy2025meta,chua2018deep} to uncertainty-aware policy search~\citep{deisenroth2011pilco,kurutach2018model,janner2019trust,frauenknecht2024trust}.
In all of these settings, the model is rolled out autoregressively over extended horizons, so predictive accuracy and well-calibrated uncertainty are not just desirable but prerequisite: small local errors compound step by step and can render predictions unreliable \citep{lambert2022investigating}.
Improving the quality of these long-horizon predictions is therefore a foundational challenge, independent of the downstream application.
Variational Bayesian last layers (VBLLs) \citep{harrison2024variational}, a specific instance of Bayesian last-layer models \cite{snoek2015scalable,riquelme2018deep,watson2021latent,fiedler2023improved}, offer a compelling architecture for this purpose, combining tractable predictive inference with the efficiency of deterministic feature extractors.
By restricting uncertainty over parameters to a Gaussian posterior over the final layer, they yield good predictive distributions at computational cost close to that of standard neural networks, and have been successfully deployed~%
\citep{lew2022safe,brunzema2025bayesian,levy2025meta,davydov2025first,brunzema2026vision}. %
Rather than arguing that VBLLs are the best probabilistic dynamics model in general, we ask how this increasingly used model class should be \emph{trained} when deployed autoregressively.

In practice, VBLLs are trained using single-step likelihood objectives, scoring predictions on isolated transitions under a Gaussian log-likelihood. 
While this objective is convenient, it implicitly fixes two design choices that are in fact independent of the variational last-layer construction itself.
First, the model is evaluated only on isolated transitions, so the learned features are shaped by single-step mini-batches rather than trajectory structure, and the model is never exposed to the compounding errors that arise in autoregressive rollouts at deployment.
Second, the data-fit term is restricted to a Gaussian log-likelihood, whose quadratic tails can be sensitive to outliers \citep{selten1998axiomatic}.
Crucially, these are limitations of the training \emph{objective}, not the \emph{model}: the Gaussian posterior over the last layer and the scoring mechanism are mathematically independent, and the VBLL architecture is compatible with a much broader class of training losses.
 
\begin{figure}[t]
    \centering
    \includegraphics[width=\linewidth]{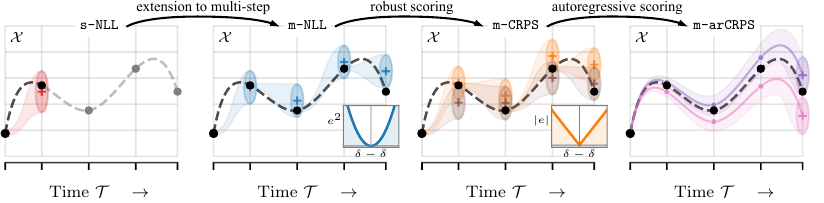}%
    \caption{\textit{Overview of the developed loss functions for VBLLs under the Gibbs VI framework.} We extend existing single-step (\selbo) to a multi-step (\melbo), robust (\mfcrps), autoregressive loss (\marcrps).}%
    \vspace{-1em}
    \label{fig:loss_overview}
\end{figure}

We develop this perspective by revisiting VBLLs through the lens of Gibbs variational inference (Gibbs VI) \citep{alquier2016properties, bissiri2016general, knoblauch2022optimization}, which generalizes standard variational Bayes by replacing the log-likelihood with an arbitrary loss.
Applied to last-layer models, this yields what we call \textit{Gibbs variational last layers}: a framework in which the training objective becomes a design choice that can encode multi-step structure, robust scoring rules, and exposure to autoregressive errors (Fig.~\ref{fig:loss_overview})---all without modifying the model architecture, the posterior family, or the tractability of predictive inference.
Within this framework, we develop a family of multi-step training objectives
that improve autoregressive performance across diverse metrics, including mean rollout approximation, distributional match, and coverage of true rollouts. %
Across a wide range of systems, we find that task-aligned training consistently improves long-horizon accuracy and uncertainty estimation, while retaining the simplicity and predictive tractability of VBLLs.

\vspace{-4pt}
\paragraph{Contributions}
In summary, our core contributions are as follows:
\begin{enumerate}[label=(\roman*), itemsep=0pt, topsep=-3pt, parsep=0pt, partopsep=0pt, leftmargin=*] %
    \item[\textbf{C1}] We introduce and formalize Gibbs variational last layers, showing that Gibbs VI strictly generalizes the standard VBLL objective and provides a principled framework for incorporating task-aligned losses into (variational) Bayesian last-layer models.
    \item[\textbf{C2}] We derive a hierarchy of multi-step training objectives within Gibbs VI that progressively introduces multi-step structure, robustness to outliers, and exposure to autoregressive errors.
    \item[\textbf{C3}] We confirm the superior performance across 18 dynamical systems and demonstrate on data from a real-world unstable robotic system that a task-aligned objective is essential when idealized assumptions break. Further, in an illustrative example, we isolate a sampling-bias failure mode of single-step likelihood training that multi-step objectives mitigate.
\end{enumerate} 

\section{Problem Setting}
\label{sec:setting}

Consider a system with state $\vx_t \in \R^{d_x}$ evolving according to the unknown stochastic transition $\vx_{t+1} = f(\vx_t) + \bm{\nu}_t$, $\bm{\nu}_t \sim p_\nu$.
We denote a trajectory segment of horizon~$H$ by $\bm{\tau}^{(n)} = (\vx_0^{(n)}, \vx_1^{(n)}, \ldots, \vx_H^{(n)})$ and assume access to a dataset $\gD = \{\bm{\tau}^{(n)}\}_{n=1}^{N}$ of $N$ such segments.
We approximate the general nonlinear system with a residual model
\begin{equation}
    \label{eq:dynamics}
    \vx_{t+1}
    = \vx_t + \mW^\top \phi_\vtheta(\vx_t) + \bm{\nu}_t,
    \qquad
    \bm{\nu}_t \sim \gN(\vzero,\, \mSigma_\nu),
\end{equation}
where $\phi_\vtheta \colon \R^{d_x} \to \R^p$ is a learned (deep) feature extractor parameterized by~$\vtheta$, $\mW \in \R^{p \times d_x}$ is the last-layer weight matrix, and $\mSigma_\nu = \operatorname{diag}(\sigma_1^2, \ldots, \sigma_{d_x}^2)$ is a diagonal process-noise covariance.
The residual structure, also used, \eg in \citep{deisenroth2011pilco}, ensures that the model defaults to the identity map, aiding optimization and reflecting the prior belief that consecutive states are similar for sufficiently small~$\Delta t$.
The framework extends naturally to controlled systems by augmenting~$\vx_t$ with a control input~$\vu_t$.

\paragraph{Variational Bayesian Last Layer}
We place an isotropic Gaussian prior on the columns $\vw_j \in \R^p$ of~$\mW$ and approximate the posterior with a mean-field Gaussian factorized across columns as in the original VBLL paper~\citep{harrison2024variational} as well as in application papers of VBLLs~\citep{brunzema2025bayesian,brunzema2026vision}:
\begin{equation}
    \label{eq:prior_posterior}
    \underbracket{p(\mW) = \textstyle\prod_{j=1}^{d_x}
    \gN(\vw_j;\, \vzero,\, \alpha\, \mI_p)}_{\text{Prior}},
    \qquad \text{and} \qquad
    \underbracket{q(\mW \mid \bm{\eta})
    = \textstyle\prod_{j=1}^{d_x}
      \gN(\vw_j;\, \vmu_j,\, \mL_j \mL_j^\top)
    }_{\text{Variational Posterior}},
\end{equation}
where $\bm{\eta} = \{(\vmu_j, \mL_j)\}_{j=1}^{d_x}$ collects the variational parameters and each $\mL_j \in \R^{p \times p}$ is a lower-triangular Cholesky factor (see Appendix~\ref{sec:implementation_vblls} for the concrete VBLL parameterization).  
Restricting uncertainty to the last layer keeps predictive inference tractable and has demonstrated strong performance in various decision making under uncertainty settings~\citep{harrison2018meta,lew2022safe,levy2025meta,de2026goal,davydov2025first,brunzema2025bayesian,brunzema2026vision,riquelme2018deep}. %

\section{Gibbs Variational Last Layers}
\label{sec:gvi}
 
The variational last-layer construction of VBLLs, \ie a Gaussian posterior over $\mW$, regularized toward a Gaussian prior via a KL penalty, is agnostic to how predictive quality is measured.
The standard VBLL objective \citep{harrison2024variational} fills this slot with a single-step Gaussian log-likelihood, but this is a \emph{choice}, not a requirement.
This modularity is precisely the premise of Gibbs VI \citep{alquier2016properties,knoblauch2022optimization,bissiri2016general}, which replaces the log-likelihood in the ELBO with an arbitrary loss function while retaining the KL divergence as a complexity penalty, yielding a variational approximation to the \emph{Gibbs posterior} \citep{bissiri2016general,catoni2007pac}.
Applying this idea to VBLLs yields what we call \emph{Gibbs variational last layers}: the loss becomes a design lever that can be tailored to the deployment setting without touching the model.
 
\subsection{General Framework}
\label{sec:gvi_framework}
 
Given this decoupling, the natural question is: \emph{What loss should one use?}
The Gibbs variational objective \citep{bissiri2016general,alquier2016properties} provides the formal framework.
We seek a variational posterior within the family $\gQ$ over the last-layer weights that minimizes expected loss on the training data while remaining close to the prior:
\begin{tikzpicture}[remember picture]
 
\newcommand{\commonheight}{\vphantom{{\textstyle\sum_{n=1}^{N} \ell(\mW,\bm{\tau}^{(n)})}}}
\definecolor{tabblue}{RGB}{31,119,180}
\definecolor{taborange}{RGB}{255,127,14}
 
\node {
\begin{minipage}{0.99\linewidth}
\begin{equation}
    \label{eq:gibbs}
    \textstyle
    q^*(\mW) \in \argmin_{q \in \gQ} \left\{\quad
        \tikzmarknode{datafit}{%
        \commonheight
        \E_{q(\mW)} \left[\sum_{n=1}^{N}
        \ell \big(\mW, \bm{\tau}^{(n)}\big)\right]}
        \quad + \lambda\quad\,
        \tikzmarknode{klterm}{%
        \commonheight
        \KL\!\left(q \mid\mid p\right)}\quad
    \right\}
\end{equation}
\end{minipage}
};
 
\begin{scope}[on background layer]
 
    \node[
        fill=tabblue!10,
        rounded corners,
        inner xsep=8pt,
        inner ysep=8pt,
        fit=(datafit),
        yshift=-6pt
    ] (boxdata) {};
 
    \path (datafit.north) -- (boxdata.north);
 
    \node[
        font=\small,
        anchor=north,
        yshift=1.4em
    ] at (boxdata.south) {\textit{Data fit term}};
 
    \node[
        fill=taborange!10,
        rounded corners,
        inner xsep=8pt,
        inner ysep=10pt,
        fit=(klterm),
        yshift=-8pt
    ] (boxkl) {};
 
    \path (klterm.north) -- (boxkl.north);
 
    \node[
        font=\small,
        anchor=north,
        yshift=1.4em
    ] at (boxkl.south) {\textit{Regularizer}};
 
\end{scope}
 
\end{tikzpicture}
where $\lambda > 0$ controls regularization strength.
The unconstrained minimizer of~\eqref{eq:gibbs} is the \emph{Gibbs posterior}~\citep{bissiri2016general,catoni2007pac} and \eqref{eq:gibbs} defines its variational approximation in $\gQ$.

The key observation is that $\ell$ is a \emph{free parameter}: when it is the negative log-likelihood, \eqref{eq:gibbs} recovers the standard VBLL objective~\citep{harrison2024variational}.
For other choices, the optimum retains a PAC-Bayesian interpretation as a trade-off between empirical risk and prior deviation~\citep{catoni2007pac,knoblauch2022optimization}.
This flexibility comes at the cost of a likelihood-based Bayesian interpretation as the resulting posterior is no longer an approximation to the conventional Bayesian posterior, and $\lambda$ becomes a tunable regularization parameter, as in tempered posterior approaches.
Nevertheless, the standard ELBO represents only one choice within a broader space of objectives and, as we will show, is not always the most effective.

In practice, we do not fix the feature extractor~$\phi_\vtheta$ but optimize it jointly with the variational parameters~$\bm{\eta}$.
We therefore seek to solve the following optimization problem:
\begin{equation}
    \label{eq:general_obj}
    (\bm{\eta}^*, \vtheta^*, \mSigma_\nu^*)
    \in
    \argmin_{\bm{\eta},\, \vtheta, \mSigma_\nu}
    \left\{
        \E_{q(\mW \mid \bm{\eta})}\!\left[
            \frac{1}{N}\sum_{n=1}^{N}
            \ell \big(\mW, \bm{\tau}^{(n)}\big)
        \right]
        + \lambda \, \KL\!\big(q(\mW \mid \bm{\eta}) \,\|\, p(\mW)\big)
    \right\}.
\end{equation}
Joint optimization is essential: while~\eqref{eq:gibbs} determines $q^*$ for a given feature map, optimizing $\vtheta$ simultaneously allows the representation to co-adapt to the choice of loss.
We will show that even for objectives that are equivalent for fixed features (\cf Corollary~\ref{cor:selbo_melbo_equiv}), joint optimization can lead to qualitatively different representations.
 
Since both $q$ and $p$ are Gaussian, the KL term admits a closed-form expression.
In the following discussions as well as the experiments, we will always use the same model parameterization with the exact same initialization to isolate the influence of the loss function on the performance.
Note that also for all loss functions, the one-step predictive distribution will remain Gaussian (Appendix~\ref{sec:app_selbo}).
 
We now instantiate~\eqref{eq:general_obj} with concrete losses that span two design axes: the \emph{scoring rule} (Gaussian log-likelihood or CRPS) and the \emph{evaluation mode} (teacher-forced or autoregressive).
Fig.~\ref{fig:loss_overview} summarizes the resulting design space that we explore in this paper.

\subsection{Scoring Rule Axis: From Log-Likelihood to CRPS}
\label{sec:likelihood_objectives}
 
\paragraph{Recovering the standard ELBO (\selbo{}).}
The simplest instantiation of~\eqref{eq:general_obj} uses the Gaussian negative log-likelihood on one-step transitions as the loss $\ell$.
With $H=1$ and $\lambda = 1/N$, this directly recovers the standard ELBO for VBLLs~\citep{harrison2024variational} (\cf Appendix~\ref{sec:app_selbo}), to which we refer as \selbo{}.
This baseline anchors the design space: every other objective we develop can be understood as relaxing one or both of the choices that \selbo{} imposes.
 
\paragraph{Extending to multi-step scoring (\melbo{}).}
The standard \selbo{} trains on isolated transitions, but at deployment the model is rolled out over many steps.
This creates a mismatch: the posterior over~$\mW$ is shaped by features at individual training states, yet never exposed to the features it will encounter along extended trajectories.
A natural extension is to change the training horizon from a single step to~$H$ steps, scoring the model's predictions along entire trajectory segments.
 
We evaluate the model in a \emph{teacher-forced} fashion: at each step~$h$, the prediction is conditioned on the true state~$\vx^{(n)}_h$.
With this, the one-step conditional distribution is
\begin{equation}
    \label{eq:teacher_forced_conditional}
    p\big(\vx^{(n)}_{h+1} \mid \vx^{(n)}_h, \mW\big)
    =
    \gN\!\big(\vx^{(n)}_{h+1};\;
    \vx^{(n)}_h + \mW^{\!\top}\phi_\vtheta(\vx^{(n)}_h),\;
    \mSigma_\nu\big).
\end{equation}
Because the conditioning state is always the ground-truth observation, successive transitions are conditionally independent given~$\mW$. The teacher-forced trajectory log-likelihood factorizes as
\begin{equation}
    \label{eq:tf_factorization}
    \textstyle
    p \big(\vx^{(n)}_{1:H} \mid \vx^{(n)}_0, \mW\big)
    \!=\!
    \prod_{h=0}^{H-1}
    p\big(\vx^{(n)}_{h+1} \mid \vx^{(n)}_h, \mW\big)
    \!=\!
    \prod_{h=0}^{H-1}
    \gN \big(
        \bm{\delta}^{(n)}_h; \;
        \mW^{\!\top}\phi_\vtheta(\vx^{(n)}_h),
        \mSigma_\nu
    \big),
\end{equation}
where $\bm{\delta}^{(n)}_h \defeq \vx^{(n)}_{h+1} - \vx^{(n)}_h$.
The \melbo{} loss directly follows as
\begin{equation}
    \label{eq:ms_nll}
    \textstyle
    \ell_{\melbo}(\mW, \bm{\tau}^{(n)})
    \;=\; -\sum_{h=0}^{H-1}
      \log \gN\!\big(
          \bm{\delta}^{(n)}_h;\;
          \mW^{\top} \phi_\vtheta(\vx^{(n)}_h),\;
          \mSigma_\nu
      \big).
\end{equation}
 
\paragraph{Closed-form expected loss.}
Since each summand in~\eqref{eq:ms_nll} is quadratic in~$\mW$, the closed-form formulation of the expected log-likelihood of standard VBLLs extends as a sum over the $H$~steps.
Stacking variational means into $\bm{\mu}_W = [\bm{\mu}_{w_1}, \ldots, \bm{\mu}_{w_{d_x}}] \in \R^{p \times d_x}$, the expectation is tractable as
\begin{equation}
    \label{eq:ms_elbo_closed}
    \E_{\viposterior}\!\big[\ell_{\melbo}(\mW, \bm{\tau}^{(n)})\big]
    \!=\!
    \sum_{h=0}^{H-1}\!
    \left(
        \!-\log \gN\!\big(\bm{\delta}^{(n)}_h; \bm{\mu}_W^{\!\top}\bm{\phi}^{(n)}_h, \mSigma_\nu\big)
        \!+\!
        \frac{1}{2}\sum_{j=1}^{d_x}
        \frac{\bm{\phi}^{(n)\top}_h \mSigma_{w_j}\, \bm{\phi}^{(n)}_h}{\sigma_j^2}\!
    \right)
\end{equation}
where $\bm{\phi}^{(n)}_h = \phi_\vtheta(\vx^{(n)}_h)$.
Therefore, as with \selbo{}, the \melbo{} objective can be optimized without Monte Carlo estimation of the expected loss.
 
\paragraph{Multi-step training reshapes features but not the objective.}
Because the teacher-forced multi-step loss~\eqref{eq:ms_nll} is simply a sum of $H$ single-step losses, the following equivalence is immediate.
 
\begin{corollary}[Equivalence of \selbo{} and \melbo{} for fixed features]
\label{cor:selbo_melbo_equiv}
For fixed $\phi_\vtheta$, if $\lambda_{\mathrm{M}} = H\,\lambda_{\mathrm{S}}$ then
$\gL_{\melbo}(\bm{\eta}) = H \cdot \gL_{\selbo}(\bm{\eta})$
and the two objectives share the same set of minimizers.
\end{corollary}
 
When $\phi_\vtheta$ is learned jointly, \melbo{} propagates gradients through features evaluated at multiple states along each trajectory segment, whereas \selbo{} extracts signal from mini-batches of one-step transitions. Although the teacher-forced likelihood still factorizes into single-step terms, this different organization of the training data can lead to different learned features (\cf Appendix~\ref{sec:different_features}). Moreover, as demonstrated by \citet{lambert2021learning}, multi-step training can increase the effective amount of supervision. Here, we leverage this by reusing overlapping trajectory segments with stride $s$, analogous to efficiency gains obtained from overlapping windows in spectral estimation \citep{welch1967use,stoica2005spectral}.

Still, the corollary also hints that the gains from moving to the multi-step log-likelihood over the single-step version are inherently limited.
The Gibbs VI framework offers a second, orthogonal lever: the scoring rule itself.
Once the loss is no longer tied to likelihood, the question becomes what properties a desirable scoring rule should have.
For long-horizon dynamics from real-world systems, a suitable scoring rule should evaluate full predictive distributions, remain robust to occasional large deviations, and admit tractable computation under Gaussian predictives.
 
\paragraph{The continuous ranked probability score.}
The CRPS~\citep{matheson1976scoring,gneiting2007strictly} satisfies these requirements.
It is a strictly proper scoring rule that rewards both calibration and sharpness, but penalizes large deviations linearly.
Because it is not a log-likelihood, it falls outside classical Bayesian inference; the Gibbs VI framework~\eqref{eq:gibbs} is precisely what allows us to use it as a principled data-fit term.
 
For a univariate random variable $X \sim P$ and observation $y \in \mathbb{R}$, the CRPS admits the energy form
\begin{equation}
    \label{eq:crps_def}
    \operatorname{CRPS}(P, y) = \E_{X\sim P}\left[|X - y| \right] - \tfrac{1}{2}\E_{X, X' \sim P} \left[ |X - X'| \right].
\end{equation}
Intuitively, the first term rewards accurate predictions, with a linear tail that limits the influence of large errors compared to the log score, while the second term rewards spread between samples and thus counteracts collapse while excessive spread is in turn penalized by the first term.
Specifically, two properties make the CRPS attractive for dynamics models: it penalizes errors approximately linearly rather than quadratically in the tails, conferring robustness to outliers; and it admits a closed form for Gaussian predictives~\citep{gneiting2005calibrated,grimit2006continuous,gneiting2007strictly} (derived in Appendix~\ref{app:crps_derivation}):
\begin{equation}
    \label{eq:crps_gaussian}
    \operatorname{CRPS} \big(\gN(\mu, \sigma^2),\, y\big)
    =
    \sigma \left[
    z\big(2\Phi(z) - 1\big) + 2\,\varphi(z) - \frac{1}{\sqrt{\pi}}
    \right],
    \qquad
    z = \frac{y - \mu}{\sigma},
\end{equation}
where $\varphi$ and $\Phi$ denote the standard normal density and CDF.
From these properties, we obtain an additional useful feature: the CRPS has gradients with respect to the location and scale parameters that are uniformly bounded as functions of the standardized residual $z$.
Applying the chain rule with $\varphi'(z) = -z\,\varphi(z)$, we obtain
\begin{alignat}{4}
    \label{eq:grad_mu_comparison}
    \frac{\partial\, \ell_{\mathrm{NLL}}}{\partial \mu}
    &= -\frac{z}{\sigma}
    &&\xrightarrow{|z| \to \infty} \;\pm\infty,
    &\qquad
    \frac{\partial\, \operatorname{CRPS}}{\partial \mu}
    &= 1 - 2\Phi(z)
    \;&&\in (-1, 1),
    \\[6pt]
    \label{eq:grad_sigma_comparison}
    \frac{\partial\, \ell_{\mathrm{NLL}}}{\partial \sigma}
    &= -\frac{z^2}{\sigma} + \frac{1}{\sigma} \;\;
    &&\xrightarrow{|z| \to \infty} \;-\infty,
    &\qquad
    \frac{\partial\, \operatorname{CRPS}}{\partial \sigma}
    &= 2\varphi(z) - \frac{1}{\sqrt{\pi}}
    \;&&\in \Big[{-}\tfrac{1}{\sqrt{\pi}},\, \tfrac{\sqrt{2}-1}{\sqrt{\pi}}\Big].
\end{alignat}
Under the NLL, a single large residual can dominate a gradient step and, via the $-z^2/\sigma$ term, inflate noise estimates to accommodate outliers.
Under the CRPS, the same residual produces a bounded gradient on both $\mu$ and $\sigma$.
These bounds hold at the score level and do not bound gradients through the backbone and under autoregressive evaluation, backpropagation through time additionally multiplies rollout Jacobians.
Empirically, we show that CRPS attenuates gradient growth, enough to keep autoregressive training finite across all tested systems and horizons, whereas training with the log score can result in gradient norms that are orders of magnitude larger (\cf, Sec.~\ref{sec:grad_norms}).

\subsection{Evaluation Mode Axis: From Teacher Forcing to Autoregressive Rollouts}
\label{sec:eval_mode}

Having introduced the scoring-rule axis, we now turn to \emph{where} predictions are evaluated during training. 
The log-likelihood losses above use teacher forcing, conditioning on ground-truth states. 
We first combine this evaluation mode with the CRPS, and then move to autoregressive evaluation, where predictions are scored along the model’s own rollout.
 
\subsubsection{Teacher-Forced CRPS (\mfcrps{})}
\label{sec:tf_crps}
 
Under teacher forcing, the predictive distribution at each step remains Gaussian given the factorization from \eqref{eq:tf_factorization}.
For trajectory~$n$, step~$h$, and output~$j$, we define
$
    P^{(n)}_{h,j}(\mW)
    =
    \gN \big(
    \hat{\delta}_{h,j}(\mW),\;
    \sigma_j^2
    \big)
$.
Applying the Gaussian CRPS~\eqref{eq:crps_gaussian} to each marginal yields the teacher-forced CRPS loss
\begin{equation}
\label{eq:fcrps_loss_clean}
\textstyle
\ell_{\mfcrps}(\mW, \bm{\tau}^{(n)})
=
\sum_{h=0}^{H-1} \sum_{j=1}^{d_x}
\operatorname{CRPS} \big(
P^{(n)}_{h,j}(\mW),\;
\delta^{(n)}_{h,j}
\big).
\end{equation}
Thus, \mfcrps{} preserves the teacher-forced structure of \melbo{} but replaces the tail-sensitive log-score with the CRPS, making the objective less dominated by occasional large residuals while still scoring calibrated and sharp predictive distributions.
 
Unlike the \melbo{}, the expected loss $\E_{q}[\ell_{\mfcrps}]$ does not admit a closed-form expression under the Gaussian posterior.
We therefore estimate it via Monte Carlo using reparameterized sampling of~$\mW$ with $S$ samples.
Note that even though the training scheme is still teacher-forced, one draw of $\mW$ is shared across and scored against the whole trajectory, providing informative multi-step gradients to both the feature extractor and the variational posterior.
 
\paragraph{Summary of teacher-forced losses.}
The two teacher-forced losses share the same structure, differing only in the scoring rule applied at each step.
Following \citet{gneiting2007strictly}, we write $\operatorname{LogS}(P, y) = -\log p(y)$ for the logarithmic score of a distribution $P$ with density $p$ and have
\begin{equation}
    \label{eq:tf_losses}
    \underbracket{
    \ell_{\melbo}^{(\mW, \bm{\tau}^{(n)})}
    =\! \sum_{h=0}^{H-1} \sum_{j=1}^{d_x}
    \operatorname{LogS} \!\big(
    P^{(n)}_{h,j}(\mW),
    \delta^{(n)}_{h,j}
    \big)
    }_{\text{Teacher-forced NLL (\melbo)}}
    \quad
    \underbracket{
    \ell_{\mfcrps}^{(\mW, \bm{\tau}^{(n)})}
    = \! \sum_{h=0}^{H-1} \sum_{j=1}^{d_x}
    \operatorname{CRPS} \!\big(
    P^{(n)}_{h,j}(\mW),
    \delta^{(n)}_{h,j}
    \big)
    }_{\text{Teacher-forced CRPS (\mfcrps)}}
\end{equation}

\subsubsection{Autoregressive Evaluation}
\label{sec:ar_crps}
 
Teacher-forced objectives, regardless of scoring rule, evaluate predictions at ground-truth states and therefore do not expose the model to the distribution shift induced by its own errors \citep{lambert2022investigating}.
At deployment, however, predictions are generated autoregressively.
To close this gap, we construct losses that evaluate the model under its own rollout dynamics.
For a fixed $\mW$, the generative dynamics remain stochastic due to process noise at every step.
Directly evaluating the resulting multi-step predictive distribution is generally intractable for nonlinear feature maps, as the distribution becomes non-Gaussian.
We therefore adopt a tractable approximation yielding a simple and effective objective that captures compounding prediction errors while retaining a closed-form scoring rule.
 
\paragraph{Deterministic rollout.}
For a fixed draw $\mW \sim q(\mW)$ and initial condition $\hat{\vx}_0(\mW) = \vx^{(n)}_0$, define the deterministic rollout
\begin{equation}
    \label{eq:ar_dynamics_rewrite}
    \hat{\vx}_{h+1}(\mW)
    \;=\;
    \hat{\vx}_h(\mW)
    + \mW^\top \phi_\vtheta\!\big(\hat{\vx}_h(\mW)\big),
    \qquad h = 0, \dots, H{-}1.
\end{equation}
This recursion defines the autoregressive mean trajectory induced by the model.
Because $\hat{\vx}_h(\mW)$ depends on previous predictions, it captures the compounding errors that teacher forcing ignores.
 
\paragraph{Gaussian predictive approximation.}
The exact predictive distribution $P_{\mW,h} = p(\vx_h \mid \mW, \vx_0)$ at horizon~$h$ is generally non-Gaussian for nonlinear feature maps, due to the repeated transformation of noise.
We seek a tractable Gaussian surrogate that is exact in the linear case and well-controlled near identity.
We approximate the predictive by decoupling mean propagation from noise accumulation:
\begin{equation}
    \label{eq:gaussian_approx_rewrite}
    P_{\mW,h}
    \;\approx\;
    \tilde{P}_{\mW,h}
    \;\defeq\;
    \gN\!\big(\hat{\vx}_h(\mW),\; h\,\mSigma_\nu\big).
\end{equation}
The mean is given by the deterministic rollout~\eqref{eq:ar_dynamics_rewrite}, while the covariance grows linearly with~$h$, reflecting cumulative process noise and neglecting its distortion by the dynamics.
A similar linearly expanding Gaussian uncertainty model has been used by~\citet{ivanovic2023expanding} for multi-step trajectory prediction under heteroskedastic noise.
The following proposition justifies this choice.

\begin{proposition}
\label{prop:linear_exactness_rewrite}
Suppose $\phi_\vtheta(\vx) = \mPhi \vx + \vb$ and define $\mA \defeq \mI + \mW^\top \mPhi$.
Then for all $h \ge 1$ the predictive distribution is Gaussian as
$
\textstyle
    p(\vx_h | \mW, \vx_0) = \gN\!\big(\hat{\vx}_h(\mW),\, \mC_h\big),
$
with
$
    \mC_h \defeq \sum_{k=0}^{h-1} \mA^k \mSigma_\nu (\mA^k)^\top.
$
We have $\mC_h = h\,\mSigma_\nu$ whenever $\mA \mSigma_\nu \mA^\top = \mSigma_\nu$ (\eg $\mA$ orthogonal and $\mSigma_\nu = \sigma^2 \mI$). Writing $\mA = \mI + \mE$, for $h\|\mE\|$ small,
$
    \|\mC_h - h\,\mSigma_\nu\| = \mathcal{O}\!\big(h^2 \|\mE\|\,\|\mSigma_\nu\|\big).
$
\hfill (Proof in Appendix~\ref{sec:proof_prop})
\end{proposition}
 
In our residual formulation $\vx_{h+1} = \vx_h + \mW^\top\phi_\vtheta(\vx_h) + \bm{\nu}_h$, the transition matrix takes the form $\mA = \mI + \mathcal{O}(\Delta t)$.
Hence, for sufficiently small time steps, the approximation operates in a near-identity regime where covariance distortion is limited.
The surrogate becomes inaccurate when the transition Jacobian strongly amplifies or contracts uncertainty, when state dimensions become strongly correlated, or when the rollout distribution becomes skewed or multimodal.
Importantly, the surrogate is only used to construct the training objective and all subsequent predictions still use the full stochastic nonlinear rollouts as in~\eqref{eq:dynamics}.
In our results, we will further show that, also with a nonlinear feature extractor, the loss remains effective; particularly on real-world data, where key assumptions such as Gaussian noise and the Markov property are likely violated.

\paragraph{Autoregressive losses.}
Since we model outputs independently, we can write \eqref{eq:gaussian_approx_rewrite} as marginals
$\tilde{P}_{\mW,h}^{(j)} = \gN\!\big(\hat{x}_{h,j}(\mW),\; h\,\sigma_j^2\big)$.
Applying the log score and the CRPS to these marginals defines: %
\begin{equation}
    \label{eq:ar_losses}
    \underbracket{
    \ell_{\marnll}^{(\mW, \bm{\tau}^{(n)})}
    = \sum_{h=1}^{H} \sum_{j=1}^{d_x}
    \operatorname{LogS} \Big(\tilde{P}_{\mW,h}^{(j)},\; x^{(n)}_{h,j} \Big)
    }_{\text{Autoregressive NLL (\marnll)}}
    \quad
    \underbracket{
    \ell_{\marcrps}^{(\mW, \bm{\tau}^{(n)})}
    = \sum_{h=1}^{H} \sum_{j=1}^{d_x}
    \operatorname{CRPS} \Big(\tilde{P}_{\mW,h}^{(j)},\; x^{(n)}_{h,j} \Big)
    }_{\text{Autoregressive CRPS (\marcrps)}}
\end{equation}
 
Both losses share the same rollout and covariance surrogate.
Under backpropagation through time, the unbounded gradients of the log score compound, which in our following experiments destabilizes \marnll on all chaotic systems, while \marcrps trains stably (\cf Appendix~\ref{sec:grad_norms}).

\paragraph{Computational considerations.}
For a fixed~$\mW$, the loss is deterministic, so stochasticity only arises from sampling $\mW \sim q(\mW)$.
With $S$ samples and horizon $H$, the forward cost is $\mathcal{O}(SH)$, as in teacher forcing.
However, unlike teacher-forced objectives, the rollout is sequential and requires backpropagation through time, making it less parallelizable and more memory-intensive in practice.

\section{Experimental Results}
\label{sec:result}
\begin{figure}[t]
    \centering
    \includegraphics[width=\textwidth]{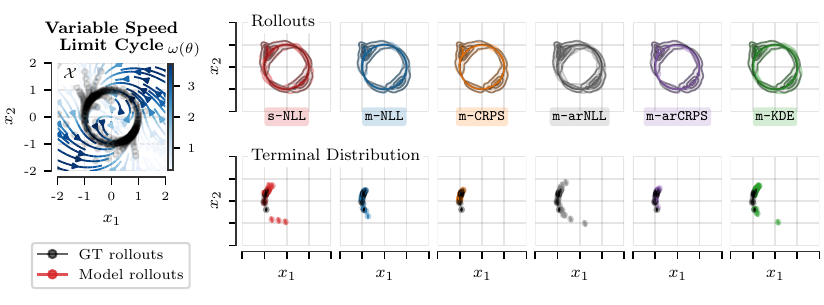}\\[-0.8em]
    \includegraphics[width=\textwidth]{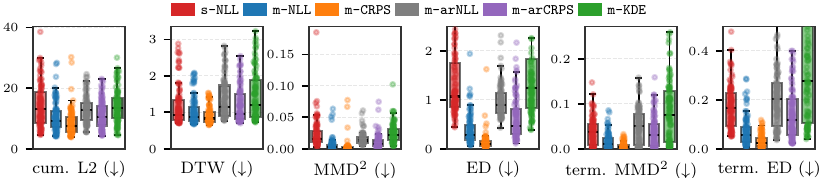}%
    \vspace{-3pt}
    \caption{\textit{Results on the VSLC.}
    \textit{Top:} Rollouts of all baselines against rollouts (color) from the true dynamics (black). 
    \textit{Bottom row:} Performance across various metrics computed from rollouts from multiple initial conditions.}%
    \vspace{-1em}
    \label{fig:vslc}
\end{figure}

We organize the evaluation into three parts.
First, we use a controlled synthetic setup to isolate a failure mode of single-step training and show how multi-step objectives can address it.
Second, we test all methods on a diverse benchmark of chaotic dynamical systems to validate results broadly.
Third, we demonstrate real-world performance on an unstable robotic system, where long-horizon prediction is particularly challenging.
We compare standard ELBO \cselbo~\citep{harrison2024variational} and autoregressive \cmkde~\citep{ivanovic2023expanding} baselines (see Sec.~\ref{sec:rw}) against teacher-forced multi-step objectives \cmelbo and \cmfcrps, and autoregressive \cmarnll and \cmarcrps.
A longer discussion on differences between \mkde, \marnll, and \marcrps in Appendix~\ref{sec:kde}.
Across experiments, we vary only the loss while keeping the model fixed to control for capacity. 
For rollouts, we sample $\mW \sim q(\mW)$ and simulate trajectories from an initial state using \eqref{eq:dynamics} with learned noise. We set $\lambda=1/|\mathcal{D}|$ and for \selbo, we use batch size $B\times H$ to ensure the same number of transitions per gradient step as multi-step methods.
Full details are provided in Appendix~\ref{sec:details} and all metrics used in the evaluations are listed in Appendix~\ref{sec:metrics}.

\subsection{Variable-Speed Limit Cycle (VSLC)}
\label{sec:results-vslc}

The VSLC is a 2D system on a globally attracting limit cycle with state-dependent angular velocity
$\omega(\theta) = \omega_0(1 + \varepsilon\sin(n_{\mathrm{modes}}\,\theta))$,
where $\theta = \mathrm{atan2}(x_2, x_1)$.
We set $\omega_0 = 2.0$, $\varepsilon = 0.85$, $n_{\mathrm{modes}} = 2$ (full dynamics in Appendix~\ref{sec:app-vslc}; flow field in Fig.~\ref{fig:vslc}, top left).

\begin{wrapfigure}{r}{0.33\textwidth}
    \centering
    \vspace{-2em}
    \includegraphics[width=\linewidth]{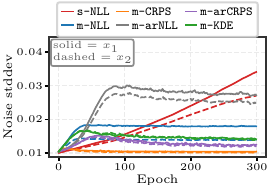}%
    \vspace{-5pt}
    \caption{\textit{Learned noise for the VSLC.} GT noise is $\sigma_1, \sigma_2 = 0.01$.}%
    \label{fig:noise_vslc}
    \vspace{-1em}
\end{wrapfigure}
Training data sampled at fixed $\Delta t$ concentrate in slow zones ($\omega(\theta) < \omega_0$), which contain ${\sim}82\%$ of samples ($F_{\mathrm{slow}} \approx 0.823$; see \Secref{sec:app-vslc}).
Single-step training thus focuses gradient signal on slow-zone states, and the larger residuals in the under-represented fast regions inflate the noise estimate (Fig.~\ref{fig:noise_vslc}).
This is a property of the log score rather than of the single-step horizon alone as the log-score-optimal variance matches the mean squared residual, so \melbo and in particular \marnll, whose autoregressive residuals grow along the rollout, inflate the noise as well.
The CRPS objectives \mfcrps and \marcrps, whose penalty grows only linearly in large residuals, recover noise estimates close to the ground truth and sample distributions matching the true distribution (Fig.~\ref{fig:vslc}).
\mkde occasionally produces divergent samples near sample boundaries (likely due to the LSE loss; \cf Appendix~\ref{sec:kde}), reflected in its larger energy distance.
Overall, \mfcrps shows the strongest performance.

\subsection{Benchmark on Chaotic Dynamical Systems}
\label{sec:results-benchmark}

\begin{wrapfigure}{r}{0.5\textwidth}
    \centering
    \vspace{-3.8em}
    \includegraphics[width=\linewidth]{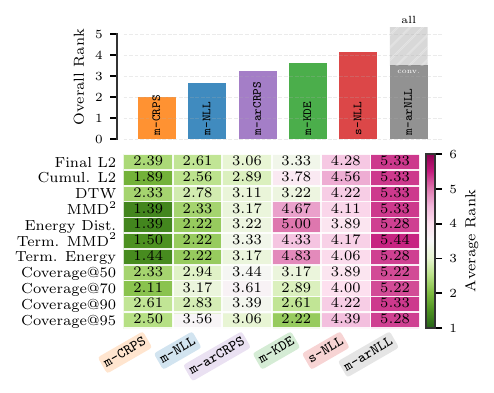}%
    \vspace{-10pt}
    \caption{\textit{Rank~summary~($\downarrow$)~across benchmark systems.}
    Full results in Tab.~\ref{tab:eval_new_v1_test_booktabs_part1} in Appendix~\ref{sec:app-additional-results}.}%
    \label{fig:results_summary}
    \vspace{-1em}
\end{wrapfigure}
After evaluating the loss functions on an environment specifically designed to highlight the potential upsides of multi-step loss function also for Markov environments, we next evaluate all methods on a broad set of benchmark of 18 dynamical systems of varying complexity, including limit cycles, an environment to test multi-modal rollouts (swimmer, \cf Appendix~\ref{sec:swimmer}), and 13 chaotic attractors from the \texttt{dysts} library~\citep{gilpin2chaos}. %
All of these are autonomous systems and satisfy the Markov property.
We generate data using rollouts under Gaussian process noise to match the setting in Sec.~\ref{sec:setting}.
\Figref{fig:results_summary} summarizes the aggregate rank across all systems and metrics.
Full per-system results are provided in \Secref{sec:app-additional-results} and Tables~\ref{tab:eval_new_v1_test_booktabs_part1}--\ref{tab:eval_new_v1_test_booktabs_part4}.

We can observe that over this large set of experiments, all stably trained multi-step loss functions outperform the single-step loss.
Similar to the VSLC, the CRPS losses outperform their likelihood-based counter parts.
\mfcrps demonstrates strong performance across the different metrics.
We can further observe that while \mkde shows strong results in terms of coverage, it struggles to match the distributions of the rollout, specifically in terms of energy distance where occasional diverging samples are strongly penalized.
This is consistent with the impressions from the VSLC results.
The matched \marnll, which differs from \marcrps only in the scoring rule, develops non-finite gradients on all chaotic systems. %
Hence, autoregressive exposure alone is not sufficient: the robustness of the CRPS is what keeps autoregressive training finite (\cf Appendix~\ref{sec:grad_norms}).

\subsection{Evaluation on Data From a Real-World Unstable Robotic System}
\label{sec:results-hardware}

\begin{figure}[t]
    \centering    
    \begin{subfigure}[b]{1.\linewidth}
        \centering
        \includegraphics[width=\linewidth]{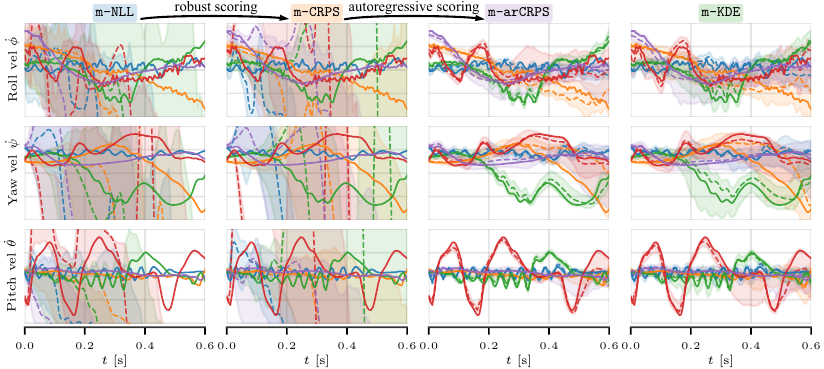}
        \vspace{-15pt}
        \caption{\textit{Autoregressive rollouts (100 steps) for the Mini Wheelbot on different trajectory segments.} 
        The solid line denotes the ground truth, the dashed line the mean rollout, and the shaded region the 5th--95th percentile range of the samples. 
        Only a subset of states is shown; the full set is provided in Fig.~\ref{fig:full_state_rollout} in the Appendix.}
    \end{subfigure}

    \begin{subfigure}[b]{0.25\linewidth}
        \centering
        \includegraphics[height=1.2in]{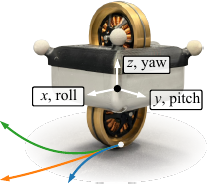}
        \caption{The Mini Wheelbot}
    \end{subfigure}
    \hfill
    \begin{subfigure}[b]{0.36\linewidth}
        \centering
        \includegraphics[width=\linewidth]{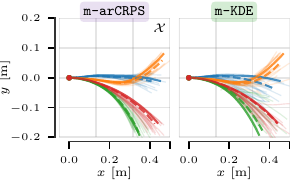}
        \vspace{-15pt}
        \caption{Rollouts in $xy$-plane (100 steps).}
    \end{subfigure}
    \hfill
    \begin{subfigure}[b]{0.36\linewidth}
        \centering
        \includegraphics[width=\linewidth]{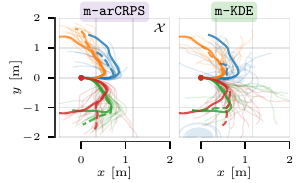}
        \vspace{-15pt}
        \caption{Rollouts in $xy$-plane (700 steps).}
    \end{subfigure}%
    \vspace{-0.5em}
    \caption{\textit{Qualitative results on data from the Mini Wheelbot.} }%
    \vspace{-1em}
    \label{fig:wb_results}
\end{figure}

We test on a dataset~\citep{hose2026mini} recorded on the Mini Wheelbot, a balancing unicycle robot~\citep{hose2025mini}.
Here, key assumptions (e.g., Gaussian noise, fully Markovian) likely break and the fast, nonlinear dynamics make modeling difficult (Appendix~\ref{app:wb}).
Table~\ref{tab:wheelbot_training_wheelbot_all_losses_v3_booktabs} reveals a clear gap: strong one-step accuracy does not imply stable multi-step rollouts.
Teacher-forced losses perform best on one-step metrics (best is \mfcrps), yet degrade or diverge over long horizons.
Figure~\ref{fig:wb_results} confirms this trend, \ie autoregressive losses are more robust.
In particular, \marcrps yields stable, accurate trajectories up to 700 steps (trained with $H=40$), while \mkde shows more chaotic, diverging behavior of some realizations.
Model-size ablations (Appendix~\ref{sec:sensitivity_model_capa}) show \marcrps consistently outperforming \mkde.
Notably, a small \marcrps model matches an \mkde model with~70$\times$~more~parameters.
Fig.~\ref{fig:sweep_over_h} further shows \marnll and \mkde diverging beyond $H=40$ and $H=60$, respectively, while \marcrps remains stable up to the maximum horizon tested of $H=100$.

\section{Related Work}
\label{sec:rw}
\paragraph{Probabilistic dynamics models.}
Probabilistic dynamics models are central to model-based control~\citep{deisenroth2011pilco,chua2018deep,janner2019trust}, and the choice of model class trades off uncertainty quality, computational cost, and ease of training.
Gaussian processes provide a principled function-space posterior and strong uncertainty estimates in the low-data regime~\citep{deisenroth2011pilco}, but generally require approximations as dataset size and dimensionality grow.
Probabilistic ensembles~\citep{lakshminarayanan2017simple} remain widely used in model-based RL~\citep{chua2018deep,janner2019trust,frauenknecht2024trust}, but they require training and evaluating multiple networks.
Fully Bayesian neural networks, \eg via MC-dropout~\citep{gal2016dropout,gal2016improving}, represent uncertainty throughout the network, but they require approximate inference over a high-dimensional parameter space. This makes them considerably more expensive, especially when training on autoregressive rollouts.
BLL models~\citep{snoek2015scalable,riquelme2018deep,watson2021latent,fiedler2023improved,harrison2024variational} offer a practical middle ground.
They share a single deterministic feature extractor and place uncertainty only over the final linear weights, which yields inexpensive epistemic uncertainty and closed-form one-step predictive moments.
Moreover, holding a weight sample fixed over a rollout corresponds to one consistent dynamics hypothesis along the trajectory.
These properties have made BLLs increasingly popular for uncertainty-aware learning and control, including online adaptation via Kalman filtering or recursive least squares~\citep{harrison2018meta,ivanovic2023expanding,levy2025meta,davydov2025first}, safe and uncertainty-aware predictive control~\citep{lew2022safe,de2026goal}, Bayesian optimization~\citep{brunzema2025bayesian}, and autonomous racing~\citep{brunzema2026vision}.
In this paper, we do not claim that BLLs dominate alternative probabilistic models in all settings, nor that they are preferable when only a mean prediction is needed; multi-step training of deterministic models is already well studied (see below).
Rather, we ask how this increasingly used model class should be trained when it is deployed autoregressively.
To this end, we build on VBLLs~\citep{harrison2024variational} but reinterpret them through Gibbs variational inference~\citep{alquier2016properties,bissiri2016general,knoblauch2022optimization}, which decouples the posterior from the training loss and enables task-aligned, multi-step objectives.

\begin{table}[tp!]
\centering
\definecolor{bestbg}{RGB}{220,235,250}
\newcommand{\best}[1]{\cellcolor{bestbg}{\boldmath $#1$}}
\caption{\textit{Quantitative results on Wheelbot data.} Multi-step metrics are reported at 100 steps.
Each cell reports $\mathrm{median}^{\mathrm{q75}}_{\mathrm{q25}}$.
Bold, shaded values are significantly best (paired Wilcoxon signed-rank test, $\alpha=0.05$).}
\label{tab:wheelbot_training_wheelbot_all_losses_v3_booktabs}
\sisetup{group-separator={,},group-minimum-digits=4,detect-weight=true,detect-inline-weight=math}
\footnotesize
\resizebox{\textwidth}{!}{%
\begin{tabular}{lcccccccc}
\toprule
 & \multicolumn{3}{c}{One-step metrics} & \multicolumn{4}{c}{Multi-step metrics} & \multicolumn{1}{c}{Training\footnotemark} \\
\cmidrule(lr){2-4}\cmidrule(lr){5-8}\cmidrule(lr){9-9}
\textbf{Method} & MSE $\downarrow$ & NLL $\downarrow$ & CRPS $\downarrow$ & Final L2 $\downarrow$ & Cumul. L2 $\downarrow$ & DTW $\downarrow$ & Energy dist. $\downarrow$ & Time [min] \\
\midrule
\texttt{s-NLL}~\citep{harrison2024variational,brunzema2026vision} & $\num{0.005}^{\num{0.019}}_{\num{0.002}}$ & $\num{-8.273}^{\num{-6.084}}_{\num{-8.932}}$ & $\num{0.328}^{\num{0.509}}_{\num{0.273}}$ & $\num{0.371}^{\num{0.759}}_{\num{0.223}}$ & $\num{12.690}^{\num{20.954}}_{\num{8.285}}$ & $\num{2.194}^{\num{3.919}}_{\num{1.161}}$ & $\num{1.909}^{\num{2.314}}_{\num{1.561}}$ & $\num{1.366}$ \\
\cmidrule(lr){1-9}
\texttt{m-NLL} & $\num{0.005}^{\num{0.019}}_{\num{0.002}}$ & $\num{-8.304}^{\num{-6.071}}_{\num{-8.982}}$ & $\num{0.340}^{\num{0.515}}_{\num{0.286}}$ & $\num{0.403}^{\num{0.722}}_{\num{0.217}}$ & $\num{13.350}^{\num{22.801}}_{\num{7.982}}$ & $\num{2.228}^{\num{3.848}}_{\num{1.089}}$ & $\num{5.181}^{\num{8.139}}_{\num{2.879}}$ & $\num{1.861}$ \\
\texttt{m-CRPS} & \best{\mathbf{0.005}^{\mathbf{0.017}}_{\mathbf{0.001}}} & \best{\mathbf{-10.676}^{\mathbf{-5.489}}_{\mathbf{-12.260}}} & \best{\mathbf{0.248}^{\mathbf{0.479}}_{\mathbf{0.169}}} & $\num{0.200}^{\num{0.259}}_{\num{0.131}}$ & $\num{7.792}^{\num{11.838}}_{\num{5.540}}$ & $\num{1.117}^{\num{1.582}}_{\num{0.767}}$ & $\num{1.456}^{\num{1.931}}_{\num{1.148}}$ & $\num{2.048}$ \\
\cmidrule(lr){1-9}
\texttt{m-KDE}~\citep{ivanovic2023expanding} & $\num{0.009}^{\num{0.034}}_{\num{0.003}}$ & $\num{-8.249}^{\num{1.104}}_{\num{-11.956}}$ & $\num{0.352}^{\num{0.707}}_{\num{0.197}}$ & $\num{0.026}^{\num{0.044}}_{\num{0.017}}$ & $\num{1.076}^{\num{1.776}}_{\num{0.678}}$ & $\num{0.087}^{\num{0.123}}_{\num{0.043}}$ & $\num{0.236}^{\num{0.285}}_{\num{0.190}}$ & $\num{3.725}$ \\
\texttt{m-arNLL} & $\num{0.010}^{\num{0.039}}_{\num{0.003}}$ & $\num{-8.011}^{\num{5.626}}_{\num{-11.677}}$ & $\num{0.363}^{\num{0.787}}_{\num{0.197}}$ & $\num{0.031}^{\num{0.045}}_{\num{0.020}}$ & $\num{1.111}^{\num{1.728}}_{\num{0.777}}$ & $\num{0.089}^{\num{0.141}}_{\num{0.053}}$ & $\num{0.223}^{\num{0.285}}_{\num{0.190}}$ & $\num{3.933}$ \\
\texttt{m-arCRPS} & $\num{0.010}^{\num{0.040}}_{\num{0.003}}$ & $\num{-0.663}^{\num{43.693}}_{\num{-12.089}}$ & $\num{0.411}^{\num{0.875}}_{\num{0.194}}$ & \best{\mathbf{0.021}^{\mathbf{0.031}}_{\mathbf{0.015}}} & \best{\mathbf{0.862}^{\mathbf{1.403}}_{\mathbf{0.573}}} & \best{\mathbf{0.065}^{\mathbf{0.099}}_{\mathbf{0.037}}} & \best{\mathbf{0.159}^{\mathbf{0.224}}_{\mathbf{0.126}}} & $\num{3.921}$ \\
\bottomrule
\end{tabular}
}
\vspace{-1em}
\end{table}

\footnotetext{All baselines were trained on the same HPC cluster, each on a single NVIDIA H100 GPU.}

\paragraph{Multi-step losses.}
Training dynamics models with multi-step losses aims to reduce compounding errors at deployment~\citep{lambert2022investigating}, and has been shown to improve long-horizon accuracy and sample efficiency in model-based RL~\citep{lambert2021learning}.
\citet{doerr2017optimizing} argue that multi-step objectives become necessary when the Markov property is violated or when process and observation noise are significant. We further show that one-step losses can fail even in controlled Markovian environments.
For Gaussian processes, moment matching and linearization-based uncertainty propagation can enable multi-step training~\citep{deisenroth2011pilco,doerr2017optimizing}. For overparameterized neural networks, an informative multi-step training signal is all the more important for learning reliable features.
For deterministic neural network models, extending the horizon yields mixed results: short horizons can help, but longer ones may degrade downstream performance~\citep{lutter2021learning,byravan2021evaluating}, which motivates careful per-step weighting~\citep{benechehab2024multi}.
For probabilistic ensembles, \citet{levy2024learning} use mean propagation over short horizons as consistent rollouts are non-trivial given the ensemble structure~\citep{frauenknecht2025rollouts}.
For VBLLs, by contrast, a consistent rollout is simply a draw of $\mW$.
Closest to our setting, \citet{ivanovic2023expanding} propose a kernel-density multi-step objective for BLLs, which we adopt as our \texttt{m-KDE} baseline. It scores a marginal mixture over sampled rollouts, whereas our objectives score each weight sample inside the Gibbs VI expectation.
Our losses build on the CRPS, a strictly proper scoring rule~\citep{matheson1976scoring,gneiting2007strictly} whose linear tail behavior makes it robust to outliers~\citep{selten1998axiomatic}.
The CRPS is widely used in weather forecasting~\citep{gneiting2007strictly,pacchiardi2024generalized,lang2026aifs}, but usually to score marginal predictive distributions directly. In this standard form it does not fit the Gibbs VI framework, since draws of $\mW$ are not scored independently.
While these individual ingredients are established, to the best of our knowledge we are the first to combine them into multi-step CRPS-based objectives for VBLLs, grounded in Gibbs VI to facilitate stable training of autoregressive dynamics models.

\paragraph{Objective mismatch and decision-aware model learning.}
Minimizing prediction error does not necessarily lead to optimal downstream task performance~\citep{lambert2020objective}, motivating decision-aware model learning approaches~\citep{wei2023unified}.
Value-aware approaches incorporate the value function into the model loss~\citep{farahmand2017value,farahmand2018iterative,voelcker2022value}, policy-aware approaches align it with the policy gradient~\citep{abachi2020policy,d2020gradient}, and goal-aware prediction conditions the model on the goal~\citep{nair2020goal}.
These methods typically operate in a deterministic setting and couple the model loss directly to policy optimization.
We share the motivation but remain in the probabilistic prediction setting, using Gibbs VI to replace the likelihood with losses aligned with how the model is deployed.
Since Gibbs VI accepts any pointwise loss, decision-aware reweightings, \eg scaling per-step CRPS terms by value-gradient magnitudes, can be incorporated directly, a promising direction for future work.
A more extensive discussion is provided in Appendix~\ref{sec:rw_appendix}.

\section{Discussion on Limitations and Future Work}
\label{sec:lims}

While our experiments demonstrate the clear benefits of objectives within the Gibbs VI framework, several limitations remain that also show promising directions for future work.

\paragraph{Calibration.}
Although the CRPS-based losses consistently improve long-horizon accuracy and distributional match, the empirical coverage of \marcrps tends to lag behind that of \mkde on several benchmark systems (\cf Table~\ref{tab:eval_new_v1_test_booktabs_part1}--\ref{tab:eval_new_v1_test_booktabs_part4}).
A key advantage of the Gibbs VI perspective is that the loss is a free design choice and calibration techniques from standard VI or tuning the regularization strength can be combined with our multi-step objectives without modifying the model or posterior family. In particular, increasing the KL weight can improve coverage (see Appendix~\ref{sec:reg_weight}), and we recommend selecting it on a held-out calibration set when coverage matters.
Going one step further, Generalized VI~\citep{knoblauch2022optimization} also generalizes the divergence enabling, \eg robust divergences~\citep{li2016renyi,futami2018variational}.
Since meaningful calibration gains can already be obtained by tuning the existing regularization, we view the KL weight as the first lever, while alternative divergences remain an interesting future direction. %

\paragraph{Empirical evaluation on downstream tasks.}
We deliberately isolated the effect of the training objective by holding the model architecture, posterior family, and optimizer fixed, attributing all observed gains unambiguously to the loss.
Consequently, we do not provide a direct comparison against other model classes such as probabilistic ensembles or GPs.
We expect downstream tasks that rely on autoregressive rollouts under learned models to benefit the most, \eg sampling-based MPC~\citep{williams2017information} or uncertainty-aware reinforcement learning~\cite{deisenroth2011pilco,janner2019trust,frauenknecht2024trust}.
We therefore view systematic evaluation in closed-loop decision-making as an important direction for future work.

\section{Conclusion}
\label{sec:conclusion}

By reinterpreting VBLLs through Gibbs variational inference, we showed that the training loss is a modular design choice—independent of the model architecture and posterior family—that can be tailored to the deployment regime.
Our hierarchy of multi-step objectives reveals two complementary axes of improvement: CRPS-based scoring confers robustness to outliers and yields superior distributional fit, while autoregressive evaluation closes the train–test mismatch that causes teacher-forced models to degrade or diverge on real-world systems.
Together, these results establish that how a VBLL is trained matters as much as what it models, opening a direct path toward integrating task-aligned losses with downstream decision-making in control and reinforcement learning.

\begin{ack}
The authors thank James Harrison, Mikkel Jordahn, John Willes, and Jasper Snoek for various insightful discussions on VBLLs over the last years.
Paul Brunzema and Henrik Hose are partially funded by the Deutsche Forschungsgemeinschaft (DFG, German Research Foundation)–RTG 2236/2 (UnRAVeL).
This work was additionally supported in part by the DFG under project number 556142469 (ParDyBO).
Simulations were performed in part with computing resources granted by RWTH Aachen University under projects rwth2066 and p0027430.
\end{ack}

\bibliography{main}  %

\newpage
\appendix

\section{Broader Impact}\label{app:impacts}
This work develops general-purpose training objectives for probabilistic dynamics models.
Improved long-horizon predictions and better-calibrated uncertainty estimates are directly relevant to safety-critical applications such as autonomous driving or robotic manipulation.
In such settings, wrong long-term prediction by models can lead to unsafe control actions, and our multi-step, CRPS-based objectives help mitigate this risk by improving distributional fit and reducing the train–test mismatch inherent in single-step training.
However, no training objective alone can guarantee safe deployment; careful system-level validation, domain-specific safety constraints, and rigorous uncertainty monitoring remain essential when integrating learned dynamics models into closed-loop control.
We do not foresee direct negative societal consequences of this work beyond those generally associated with improved automation capabilities.

\section{On the Parameterization and Implementation of VBLLs}
\label{sec:implementation_vblls}

We follow closely the original VBLL codebase of \citet{harrison2024variational} and re-implement it in JAX using Flax NNX and Optax\footnote{Flax NNX and Optax are both publicly available under a Apache-2.0 license.}.
This appendix details the concrete parameterization of the variational posterior covariance matrices and the process-noise covariance, both of which require some care to ensure that the resulting matrices remain positive definite throughout optimization.

\paragraph{Cholesky parameterization of the posterior covariance.}
As stated in \eqref{eq:prior_posterior}, the variational posterior over each column $\vw_j$ of $\mW$ is $\gN(\vmu_j, \mSigma_j)$ with $\mSigma_j = \mL_j \mL_j^\top$, where $\mL_j \in \R^{p \times p}$ is lower triangular with positive diagonal.
To enforce this structure during unconstrained gradient-based optimization, we store two sets of free parameters per output dimension~$j$: a vector $\vd_j \in \R^p$ of unconstrained log-diagonal entries and a matrix $\mO_j \in \R^{p \times p}$ of unconstrained off-diagonal entries.
The Cholesky factor is then assembled as
\begin{equation}
    \mL_j = \operatorname{tril}(\mO_j, -1) + \operatorname{diag}\!\big(\exp(\vd_j)\big),
\end{equation}
where $\operatorname{tril}(\cdot, -1)$ extracts the strictly lower-triangular part.
The exponential map on the diagonal guarantees $(\mL_j)_{kk} > 0$ for all~$k$, which in turn ensures that $\mSigma_j = \mL_j \mL_j^\top$ is positive definite for any value of the free parameters.
We initialize $\vd_j$ near $-\tfrac{1}{2}\log p$ and both $\mO_j$ and $\vmu_j$ with small random values, so that the posterior starts concentrated near zero with low uncertainty.
For all losses, we always use the same initialization.

\paragraph{Process-noise parameterization.}
The diagonal process-noise covariance $\mSigma_\nu = \operatorname{diag}(\sigma_1^2, \ldots, \sigma_{d_x}^2)$ is parameterized through unconstrained log-standard-deviations $\lambda_j \in \R$, so that $\sigma_j = \exp(\lambda_j)$.
This again ensures positivity without constrained optimization.
The parameters~$\lambda_j$ are initialized at $\log(\sigma_{\mathrm{init}})$ for a user-specified $\sigma_{\mathrm{init}} > 0$ (we use $\sigma_{\mathrm{init}} = 0.01$ by default), reflecting the prior belief that process noise is initially small.

\section{Experimental Details}
\label{sec:details}

In the following, we will list the experimental details for the empirical studies and introduce the system dynamics where necessary.
We first list general settings, \ie the model architecture and the optimizer.
All code is publicly available under the MIT license at \url{https://github.com/brunzema/gibbs-vbll}.

\paragraph{Model architecture}
As a backbone we use a standard feed forward neural network with three layers and an additional layer that maps to the features.
In Tab.~\ref{tab:cluster_training_hparams:model} and Tab.~\ref{tab:cluster_training_hparams:wheelbot_model} we list the parameters for the specific set of experiments.

\paragraph{Optimizer and training}
For all experiments, we employ the AdamW optimizer \citep{loshchilov2017decoupled} with a learning rate of $1\times10^{-3}$ ($5\times10^{-4}$ for the Mini Wheelbot experiments in Sec.~\ref{sec:results-hardware}).
We apply a weight decay of $1\times10^{-4}$ exclusively to the backbone, following standard practice for VBLLs, as the final layer is regularized through the KL term in the loss \citep{harrison2024variational}.
In addition, we use gradient clipping with a maximum norm of one in all experiments.
We train for 300 epochs with a batch size of 64 on the VSLC dataset and the chaotic systems benchmark.
For the Mini Wheelbot experiments, we train for 400 epochs with a batch size of 256.
We choose these training durations to ensure sufficient convergence of each loss function and select the best model state based on a validation set, evaluating performance every 5 epochs.
More details in Tab.~\ref{tab:cluster_training_hparams:environments} and Tab.~\ref{tab:cluster_training_hparams:wheelbot_losses}.

\subsection{Variable-Speed Limit Cycle}
\label{sec:app-vslc}

The VSLC state $\vx = (x_1, x_2)^\top \in \R^2$ evolves according to
\begin{equation}
    \dot{x}_1 = \mu(R - r)\tfrac{x_1}{r} - \omega(\theta)\,x_2, \qquad
    \dot{x}_2 = \mu(R - r)\tfrac{x_2}{r} + \omega(\theta)\,x_1,
    \label{eq:vslc-dynamics}
\end{equation}
with $r = \|\vx\|_2$ and $\theta = \mathrm{atan2}(x_2, x_1)$.
The parameter $\mu > 0$ governs radial attraction toward the limit cycle of radius $R$, while $\omega(\theta)$ controls angular motion.
Under the parameterization in Sec.~\ref{sec:results-vslc}, the angular velocity ranges over
$\omega(\theta) \in [\omega_0(1-\varepsilon),\, \omega_0(1+\varepsilon)]$,
which evaluates to $[0.30,\,3.70]\;\mathrm{rad/s}$ for the chosen parameters.
Training data collected at a fixed time step $\Delta t$ induce a distribution over angles that is proportional to the time spent at each state.
Since angular speed is $\omega(\theta)$, this yields a density proportional to $\omega(\theta)^{-1}$.

Defining the slow zone by $\omega(\theta) < \omega_0$, the fraction of samples originating from that region is
\begin{equation}
    F_{\mathrm{slow}}
    = \frac{\displaystyle\int_{\{\theta:\,\omega(\theta)<\omega_0\}} \omega(\theta)^{-1}\,d\theta}
           {\displaystyle\int_0^{2\pi} \omega(\theta)^{-1}\,d\theta}.
    \label{eq:slow-fraction}
\end{equation}
Using $\omega(\theta)=\omega_0\bigl(1+\varepsilon\sin(n_{\mathrm{modes}}\theta)\bigr)$ and the substitution $\phi = n_{\mathrm{modes}}\theta$, we obtain
\begin{equation}
    F_{\mathrm{slow}}
    =
    \frac{\displaystyle\int_{\{\phi:\,\sin\phi<0\}} \frac{d\phi}{1+\varepsilon\sin\phi}}
         {\displaystyle\int_0^{2\pi} \frac{d\phi}{1+\varepsilon\sin\phi}}.
\end{equation}
This expression is independent of $n_{\mathrm{modes}}$ (for integer $n_{\mathrm{modes}}$).
Evaluating the integrals yields the closed-form result
\begin{equation}
    F_{\mathrm{slow}} = \frac{1}{2} + \frac{\arcsin(\varepsilon)}{\pi}.
\end{equation}
For our parameter setting of $\varepsilon = 0.85$, this gives $F_{\mathrm{slow}} \approx 0.823$.
Thus, approximately $82.3\%$ of all training samples lie in the slow zone.

\paragraph{Parameters and data generation.}
We use the parameter values $\omega_0 = 2.0$, $\varepsilon = 0.85$, $n_{\mathrm{modes}} = 2$.
Trajectory data are generated by numerically integrating the system with sampling time $\Delta t = \SI{0.1}{\second}$ and induce process noise on all the states with $\sigma_1 = \sigma_2 = 0.01$.
We generate a total of 300 trajectories with each having a length of 90 time steps from pseudo-random initial conditions within $x, y \in [-2, 2]$ using a Sobol sequence.
\SI{20}{\percent} of trajectories are used as validation data for all the methods to determine the best performing model state which is then used for testing on newly random generated trajectories.
For the multi-step losses, we use a horizon of $H=40$ and a stride of one to effectively increase the supervision signal. 

\subsection{Swimmer}
\label{sec:swimmer}

\begin{figure}
    \centering
    \includegraphics[width=\linewidth]{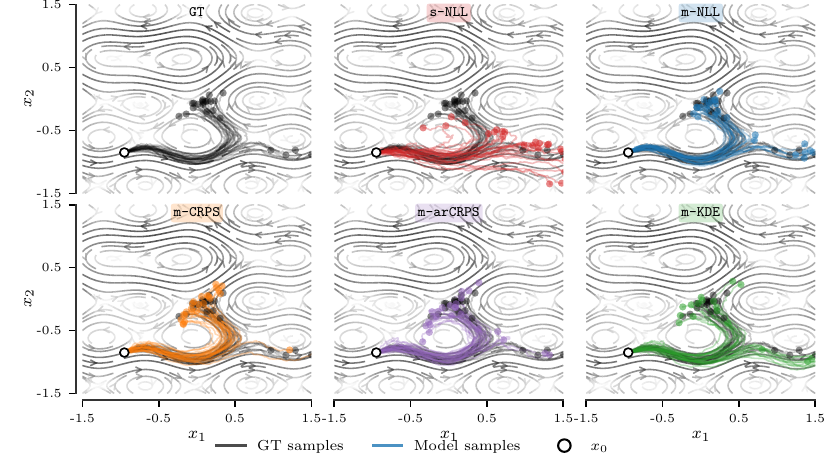}
    \caption{\textit{Model rollouts on Swimmer (flow field in gray).} We test whether the different losses allow the model to capture complex nonlinear dynamics resulting in multi-modal rollouts.
    We can see that all losses allow the model demonstrate such complex behavior.
    Still, multi-step losses allow for a better approximation of the distribution.}
    \label{fig:blabla}
\end{figure}

The Swimmer environment is a two-dimensional continuous-time dynamical system that combines a spatially varying tidal-drag field with a localized vortex flow.
The state is $\vx = (x,y)^\top \in \R^2$, and its dynamics are given by
\begin{equation}
    \dot{\vx} = \vv_{\mathrm{tidal}}(\vx) + \vv_{\mathrm{vortex}}(\vx).
    \label{eq:swimmer-dynamics}
\end{equation}

\paragraph{Tidal-drag component.}
The tidal field is defined through an oscillatory base flow
\begin{equation}
    \vc_0(\vx)
    =
    \begin{pmatrix}
        U_0 \sin(\kappa y) \\
        V_0 \sin(\kappa x)
    \end{pmatrix},
\end{equation}
with fixed amplitudes $U_0 = 0.4$ and $V_0 = 0.2$.
To attenuate the flow in regions of high instantaneous magnitude, this base field is scaled by a damping factor
\begin{equation}
    \vv_{\mathrm{tidal}}(\vx)
    =
    \frac{\vc_0(\vx)}{1 + \eta \|\vc_0(\vx)\|_2^2},
    \label{eq:swimmer-tidal}
\end{equation}
where $\eta > 0$ controls the strength of the damping.

\paragraph{Vortex component.}
The vortex is centered at
$
\vc = (-0.5,\,0.5)^\top
$
with fixed radius parameter $R_{\mathrm{v}} = 0.5$.
Writing $\vd = \vx - \vc = (d_x, d_y)^\top$, the vortex-induced velocity is
\begin{equation}
    \vv_{\mathrm{vortex}}(\vx)
    =
    \frac{\alpha R_{\mathrm{v}}^2}{\|\vd\|_2^2 + R_{\mathrm{v}}^2}
    \begin{pmatrix}
        -d_y \\
        d_x
    \end{pmatrix},
    \label{eq:swimmer-vortex}
\end{equation}
where $\alpha > 0$ sets the vortex strength.
This term induces rotational motion around the vortex center, with magnitude that decays smoothly away from the core.

\paragraph{Full dynamics.}
Combining \eqref{eq:swimmer-tidal} and \eqref{eq:swimmer-vortex}, the Cartesian dynamics are
\begin{equation}
    \dot{x}
    =
    \frac{U_0 \sin(\kappa y)}{1 + \eta \|\vc_0(\vx)\|_2^2}
    -
    \frac{\alpha R_{\mathrm{v}}^2}{\|\vx-\vc\|_2^2 + R_{\mathrm{v}}^2}(y-c_y),
\end{equation}
\begin{equation}
    \dot{y}
    =
    \frac{V_0 \sin(\kappa x)}{1 + \eta \|\vc_0(\vx)\|_2^2}
    +
    \frac{\alpha R_{\mathrm{v}}^2}{\|\vx-\vc\|_2^2 + R_{\mathrm{v}}^2}(x-c_x),
\end{equation}
where $\vc=(c_x,c_y)^\top=(-0.5,0.5)^\top$ which are clearly highly non-linear.

\paragraph{Parameters and data generation.}
We use the parameter values $\alpha = 0.7$, $\eta = 1.0$, and $\kappa = 1.5\pi$.
Trajectory data are generated by numerically integrating the system with sampling time $\Delta t = \SI{0.1}{\second}$ and induce process noise on all the states with $\sigma_1 = \sigma_2 = 0.01$.
We generate a total of 300 trajectories with each having a length of 90 time steps from pseudo-random initial conditions within $x, y \in [-1.5, 1.5]$ using a Sobol sequence.
\SI{20}{\percent} of trajectories are used as validation data for all the methods to determine the best performing model state which is then used for testing on newly random generated trajectories.
For the multi-step losses, we use a horizon of $H=40$ and a stride of one to effectively increase the supervision signal. 

\clearpage
\subsection{The Mini Wheelbot}
\label{app:wb}
\begin{wrapfigure}{r}{0.53\textwidth}
        \vspace{-2em}
        \centering
        \includegraphics[width=0.9\linewidth]{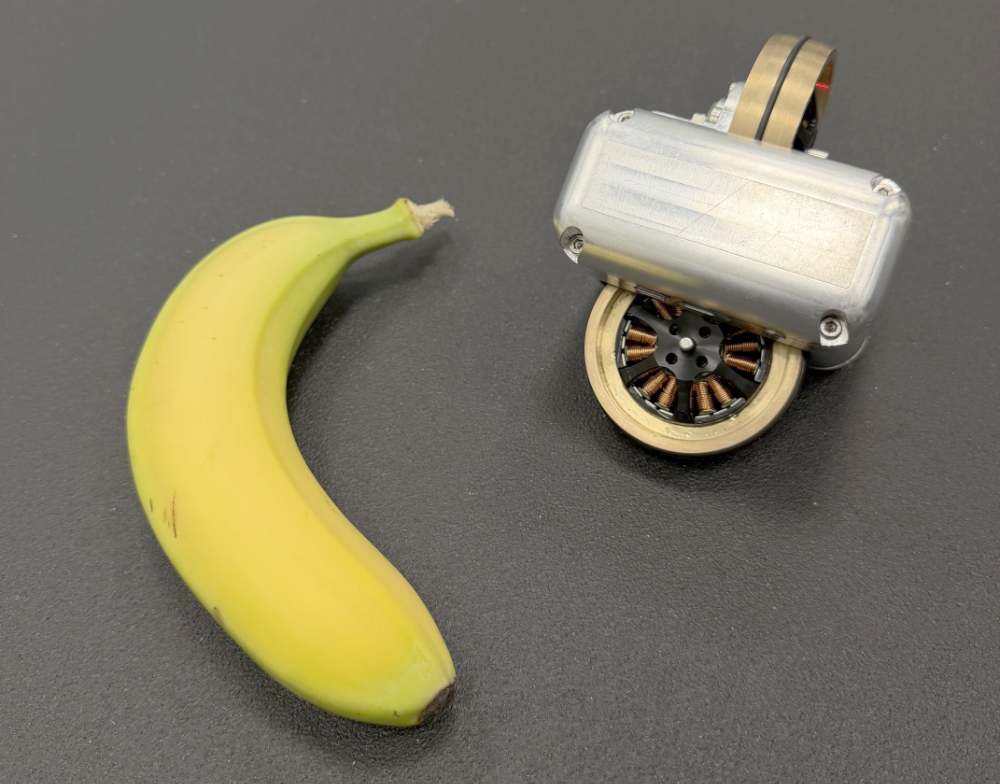}
        \caption{\textit{The Mini Wheelbot \citep{hose2025mini}.} Banana for scale.}
        \vspace{-1em}
\end{wrapfigure}
Reaction wheel balancing robots are challenging control benchmarks that have attracted research attention for multiple decades due to fast, unstable dynamics and strong nonlinearities (see~\citep{rizal2026design} for a recent survey).
The Mini Wheelbot is such an open-source robot~\citep{hose2025mini} that is based on an earlier prototype~\cite{geist2022wheelbot}.
The Mini Wheelbot uses its two wheels for balancing: the driving wheel in the pitch direction like a segway and the reaction wheel in the roll direction like a reaction wheel pendulum.
The robot has no third ``turntable'' actuator for controlling the yaw orientation.
Instead, it relies on geometric turning of a single wheel at an angle and the nonlinear gyroscopic effects of the multibody dynamics for reorienting.

\paragraph{Data and training setup} 
To train the models, we use data from the Mini Wheelbot dataset \citep{hose2026mini}\footnote{The dataset is available under the MIT license at \url{https://github.com/wheelbot/dataset}.}.
This data is collected at $\Delta t = \SI{0.006}{\second}$ and we use the same train, validation, and test split for all loss functions.
For all the trajectories, we cut the first \SI{2}{\second} and the last \SI{2}{\second}.
For training and evaluation, we replay the ground-truth actions over the trajectory segment.

\paragraph{Mapping state rollouts onto the $xy$-plane}
To visualize rollouts in the plane, we map the predicted dynamics to a planar trajectory on the $xy$-plane via unicycle-like wheel odometry.
Let $r_w$ denote the drive-wheel radius, $\Delta t$ the sampling period, $q_\mathrm{DR}$ the drive-wheel angle, and $\psi$ the yaw angle.
We first align the yaw to the world frame,
\begin{equation}
    \psi_t^{\mathrm{w}} \;=\; \mathrm{unwrap}(\psi_t) \;-\; \psi_0 \;+\; \psi^{\mathrm{w}}_\mathrm{init},
    \label{eq:wb_yaw_world}
\end{equation}
where $\mathrm{unwrap}(\cdot)$ removes $2\pi$ discontinuities and $\psi^{\mathrm{w}}_\mathrm{init}$ is a chosen initial heading (in Fig.~\ref{fig:wb_results}, we choose $\psi^{\mathrm{w}}_\mathrm{init}=0$ for all segments).
The arc length traveled by the drive wheel between two consecutive samples is
\begin{equation}
    \Delta s_t \;=\; r_w \,\bigl(q_{\mathrm{DR},t} - q_{\mathrm{DR},t-1}\bigr),
    \label{eq:wb_arc_length}
\end{equation}
and, assuming no wheel slip and motion along the current heading, the planar position is accumulated by projecting each increment onto the heading direction, starting from $(x_0, y_0)=(0,0)$:
\begin{equation}
    \begin{aligned}
        x_t &\;=\; \sum_{k=1}^{t} \cos\!\bigl(\psi_k^{\mathrm{w}}\bigr)\,\Delta s_k, &
        y_t &\;=\; \sum_{k=1}^{t} \sin\!\bigl(\psi_k^{\mathrm{w}}\bigr)\,\Delta s_k.
    \end{aligned}
    \label{eq:wb_odometry}
\end{equation}
For model rollouts, the learned dynamics output the yaw rate $\dot\psi$ and drive-wheel speed $\dot q_\mathrm{DR}$ rather than $\psi$ and $q_\mathrm{DR}$ directly, as the MW is invariant to these absolute values.
We reconstruct the latter by forward-Euler integration from the measured initial values $\psi_0, q_{\mathrm{DR},0}$,
\begin{equation}
    \begin{aligned}
        \psi_{t+1} &\;=\; \psi_t + \dot\psi_t\,\Delta t, &
        q_{\mathrm{DR},t+1} &\;=\; q_{\mathrm{DR},t} + \dot q_{\mathrm{DR},t}\,\Delta t,
    \end{aligned}
    \label{eq:wb_euler}
\end{equation}
and then apply the odometry mapping above to obtain $(x_t, y_t)$.

\newpage
\section{Metrics}
\label{sec:metrics}

Throughout, let $\hat{\vx}_{0:H} = (\hat{\vx}_0, \ldots, \hat{\vx}_H)$ denote the model's mean predicted trajectory, $\vx^*_{0:H}$ the noiseless ground-truth trajectory, $\{\tilde{\vx}^{(s)}_{0:H}\}_{s=1}^{S}$ the noisy ground-truth rollouts, and $\{\hat{\vx}^{(s)}_{0:H}\}_{s=1}^{S}$ the stochastic model rollout samples, all initialized from the same state $\vx_0 \in \R^d$.
With this we can define the following metrics of interest.

\paragraph{Final Displacement Error.}
The Final Displacement Error (FDE) measures the Euclidean distance between the mean predicted state and the noiseless ground truth at the terminal timestep $H$:
\begin{equation}
    \label{eq:fde}
    \mathrm{FDE} = \|\hat{\vx}_H - \vx^*_H\|_2.
\end{equation}

\paragraph{Cumulative L2 Error.}
The Cumulative L2 Error sums the per-step Euclidean distances between the mean predicted trajectory and the noiseless ground truth over all timesteps:
\begin{equation}
    \label{eq:cumulative_l2}
    \mathrm{CL2} = \sum_{h=1}^{H} \|\hat{\vx}_h - \vx^*_h\|_2.
\end{equation}
This measures the total accumulated trajectory deviation rather than only the terminal error.

\paragraph{Dynamic Time Warping Distance.}
Dynamic Time Warping (DTW)~\citep{sakoe1978dynamic} is a similarity measure between temporal sequences that is robust to local timing differences.
For multi-dimensional trajectories we compute the DTW distance per state dimension and sum:
\begin{equation}
    \label{eq:dtw}
    \mathrm{DTW} = \sum_{j=1}^{d} \mathrm{DTW}_j\!\left(\{\hat{x}_{h,j}\}_{h=0}^H,\, \{x^*_{h,j}\}_{h=0}^H\right),
\end{equation}
where $\mathrm{DTW}_j$ denotes the scalar DTW distance along dimension $j$.
Unlike CL2, DTW allows for non-linear temporal alignment, making it less sensitive to phase shifts in oscillatory dynamics.
To evaluate this metric, we use the Python package \texttt{dtaidistance}\footnote{\texttt{dtaidistance} is publicly available under the Apache Licence, Version 2.0 at \url{https://github.com/wannesm/dtaidistance}.} \citep{wannesm2022wannesm} for time-series distances.

\paragraph{Maximum Mean Discrepancy.}
To compare the full predictive distribution against the ground-truth distribution over trajectories, we use the unbiased estimator of the squared Maximum Mean Discrepancy (MMD$^2$)~\citep{gretton2012kernel} with an RBF kernel.
Each trajectory is flattened into a vector $\vz = \mathrm{vec}(\vx_{0:H}) \in \R^{d(H+1)}$, giving sample sets $\{\hat{\vz}^{(s)}\}_{s=1}^S$ and $\{\tilde{\vz}^{(s)}\}_{s=1}^S$ for the model and noisy ground truth respectively.
The kernel is
\begin{equation}
    k(\vz, \vz') = \exp\!\left(-\frac{\|\vz - \vz'\|_2^2}{2\sigma^2}\right),
\end{equation}
where $\sigma$ is set via the median heuristic \citep[Section~8]{gretton2012kernel} applied to the ground-truth samples and held fixed across all models. 
The unbiased MMD$^2$ estimator is
\begin{equation}
    \label{eq:mmd}
    \widehat{\mathrm{MMD}}^2 = \frac{1}{S(S-1)}\sum_{s \neq s'} k(\hat{\vz}^{(s)}, \hat{\vz}^{(s')})
    + \frac{1}{S(S-1)}\sum_{s \neq s'} k(\tilde{\vz}^{(s)}, \tilde{\vz}^{(s')})
    - \frac{2}{S^2}\sum_{s,s'} k(\hat{\vz}^{(s)}, \tilde{\vz}^{(s')}).
\end{equation}
A value of $\widehat{\mathrm{MMD}}^2 = 0$ indicates that the model and ground-truth trajectory distributions are indistinguishable under the kernel.

\paragraph{Energy Distance.}
The Energy Distance~\citep{szekely2013energy} is a strictly proper divergence between distributions that does not require a kernel bandwidth.
Using the same flattened trajectory representations as above, the empirical Energy Distance is
\begin{equation}
    \label{eq:energy_distance}
    \widehat{D}_E = 2\,\E\!\left[\|\hat{\vz} - \tilde{\vz}\|_2\right] - \E\!\left[\|\hat{\vz} - \hat{\vz}'\|_2\right] - \E\!\left[\|\tilde{\vz} - \tilde{\vz}'\|_2\right],
\end{equation}
where expectations are estimated by averaging over all pairs of samples.
The Energy Distance is equivalent to MMD$^2$ with the negative-definite kernel $k(\vz, \vz') = -\|\vz - \vz'\|_2$, and therefore operates directly in the original Euclidean trajectory space rather than in a kernel-induced feature space.
As a consequence, $\widehat{D}_E$ penalizes distributional discrepancies proportionally to Euclidean distances between trajectories: two distributions are close in Energy Distance if and only if their samples are close on average in $\ell_2$.
This makes it interpretable and scale-equivariant, but unlike the RBF kernel used in MMD$^2$, it does not discount large deviations exponentially and thus is sensitive to outlying trajectories.
Like MMD$^2$, $\widehat{D}_E = 0$ if and only if the two distributions are equal.

\paragraph{Empirical Marginal Coverage.}
Let $\{\hat{\vx}^{(s)}_{0:H}\}_{s=1}^{S}$ denote $S$ stochastic rollout samples from the model and $\{\tilde{\vx}^{(s)}_{0:H}\}_{s=1}^{S}$ denote $S$ noisy ground-truth rollouts.
For a nominal level $1 - \alpha \in (0,1)$, define the per-dimension marginal quantile bounds at each timestep $h$ and dimension $j$ as
\begin{equation}
    \hat{q}^{-}_{h,j} = Q_{\alpha/2}\!\left(\{\hat{x}^{(s)}_{h,j}\}_{s=1}^{S}\right),
    \qquad
    \hat{q}^{+}_{h,j} = Q_{1-\alpha/2}\!\left(\{\hat{x}^{(s)}_{h,j}\}_{s=1}^{S}\right),
\end{equation}
where $Q_\beta(\cdot)$ denotes the $\beta$-quantile of a set of scalars.
The empirical marginal coverage at timestep $h$ is
\begin{equation}
    \label{eq:coverage}
    \widehat{\mathrm{Cov}}(h) = \frac{1}{d} \sum_{j=1}^{d} \frac{1}{S} \sum_{s=1}^{S} \mathbf{1}\!\left[\hat{q}^{-}_{h,j} \leq \tilde{x}^{(s)}_{h,j} \leq \hat{q}^{+}_{h,j}\right].
\end{equation}
A perfectly calibrated model satisfies $\widehat{\mathrm{Cov}}(h) \approx 1 - \alpha$ for all $h$.
We report the time-averaged marginal coverage, excluding the trivial $h=0$ step at which all samples coincide with the known initial condition:
\begin{equation}
    \label{eq:mean_coverage}
    \overline{\mathrm{Cov}} = \frac{1}{N} \sum_{i=1}^{N} \frac{1}{H} \sum_{h=1}^{H} \widehat{\mathrm{Cov}}_i(h).
\end{equation}

\begin{remark}
    We compute coverage per dimension and then average, rather than checking whether all dimensions are simultaneously inside their respective intervals (joint coverage).
    To see why this matters, suppose each per-dimension marginal interval is correctly calibrated at nominal level $1-\alpha$. 
    The probability that all $d$ dimensions are \emph{simultaneously} inside their intervals is bounded by
    \begin{equation}
        \max(0,\, 1-d\alpha) \leq P\!\left(\bigcap_{j=1}^{d} \hat{x}^{(j)} \in [\hat{q}^{-}_j,\, \hat{q}^{+}_j]\right) \leq 1-\alpha,
    \end{equation}
    and equals $(1-\alpha)^d$ under independence across dimensions. 
    For a $d=2$ system with independent dimensions this gives a joint coverage of $(0.90)^2 = 0.81$ at the $90\%$ nominal level and $(0.95)^2 = 0.90$ at $95\%$, even for a perfectly calibrated model. 
    Reporting joint coverage against a nominal level of $1-\alpha$ would therefore understate calibration for any $d > 1$ unless the dimensions are perfectly dependent, with the gap growing exponentially in $d$ under independence. 
    Averaging marginal coverage across dimensions as in~\eqref{eq:coverage} removes this dependence on $d$ and yields a quantity that is directly comparable to the nominal level regardless of state dimension.
\end{remark}

\newpage
\section{Details on Baselines}

We first show that the one-step predictive distribution is independent of the training loss, then state the baselines not derived in the main text.

\paragraph{Loss-independent one-step predictive distribution.}
For a test point $\vx_h$, the one-step predictive is obtained by marginalizing over the posterior:
\begin{equation}
    p(\vx_{h+1} \mid \vx_h)
    = \int p(\vx_{h+1} \mid \vx_h, \mW)\, q(\mW \mid \bm{\eta})\, d\mW.
\end{equation}
The likelihood $p(\vx_{h+1} \mid \vx_h, \mW) = \gN(\vx_{h+1};\, \vx_h + \mW^\top \phi_\vtheta(\vx_h),\, \mSigma_\nu)$ is linear in~$\mW$ for fixed features $\bm{\phi}_h = \phi_\vtheta(\vx_h)$, and $q(\mW \mid \bm{\eta})$ is Gaussian by construction of the variational family~$\gQ$.
Since a Gaussian integrated against a linear-Gaussian observation model yields a Gaussian, the predictive is available in closed form.
Concretely, because the outputs are modeled independently with $q(\vw_j) = \gN(\vmu_j, \mL_j \mL_j^\top)$, the predicted residual for dimension~$j$ satisfies
\begin{equation}
    \E_q[\vw_j^\top \bm{\phi}_h] = \vmu_j^\top \bm{\phi}_h,
    \qquad
    \operatorname{Var}_q[\vw_j^\top \bm{\phi}_h] = \bm{\phi}_h^\top \mL_j \mL_j^\top \bm{\phi}_h.
\end{equation}
Adding the aleatoric noise~$\sigma_j^2$, the marginal predictive for each output dimension is
\begin{equation}
    \label{eq:predictive_closed_form}
    p(\delta_{h,j} \mid \vx_h)
    = \gN\!\big(\vmu_j^\top \bm{\phi}_h,\;
    \sigma_j^2 + \bm{\phi}_h^\top \mL_j \mL_j^\top \bm{\phi}_h\big).
\end{equation}
The predictive contains variational parameters~$\bm{\eta}$, the noise~$\sigma_j^2$, and the features~$\bm{\phi}_h$.
The choice of scoring rule and evaluation mode affects the \emph{values} of the optimized parameters $(\bm{\eta}^*, \vtheta^*, \mSigma_\nu^*)$, but the \emph{functional form} of the predictive remains an axis-aligned Gaussian.

\subsection{Single-Step ELBO (\selbo{})}
\label{sec:app_selbo}

For completeness we state the single-step special case of the \melbo{} objective
(Section~\ref{sec:likelihood_objectives}) explicitly.
Setting $H=1$ (and therefore inducing a dataset of one-step transitions, \ie $\gD = \{(\vx^{(n)}_0, \vx^{(n)}_1)\}_{n=1}^{N}$), the loss reduces to the Gaussian negative log-likelihood of a single transition:
\begin{equation}
    \label{eq:ss_nll}
    \underbracket{\ell_{\selbo}(\mW, \bm{\tau}^{(n)})
    \;=\; -\log \gN\!\big(
          \bm{\delta}^{(n)}_0;\;
          \mW^{\top} \phi_\vtheta(\vx^{(n)}_0),\;
          \mSigma_\nu
      \big),
      }_{\text{Single-step NLL (\selbo)}}
\end{equation}
where $\bm{\delta}^{(n)}_0 \defeq \vx^{(n)}_{1} - \vx^{(n)}_0$.
Writing $\bm{\phi}_n = \phi_\vtheta(\vx^{(n)}_0)$ and stacking variational means into
$\bm{\mu}_W = [\bm{\mu}_{w_1}, \ldots, \bm{\mu}_{w_{d_x}}] \in \R^{p \times d_x}$,
the closed-form expected loss under the Gaussian posterior is
\begin{equation}
    \label{eq:ss_elbo_closed}
    \E_{\viposterior}\!\big[\ell_{\selbo}(\mW, \bm{\tau}^{(n)})\big]
    \;=\;
    -\log \gN\!\big(\bm{\delta}^{(n)}_0;\; \bm{\mu}_W^{\!\top}\bm{\phi}_n,\; \mSigma_\nu\big)
    \;+\;
    \frac{1}{2}\sum_{j=1}^{d_x} \frac{\bm{\phi}_n^{\!\top}\mSigma_{w_j}\bm{\phi}_n}{\sigma_j^2},
\end{equation}
With $\lambda = 1/N$ in \eqref{eq:gibbs}, this recovers the standard VBLL of~\citet{harrison2024variational}.

\subsection{Kernel Density Estimation of the Marginal Likelihood (\mkde)}
\label{sec:kde}

\begin{wrapfigure}{r}{0.25\textwidth}
        \vspace{-2em}
        \centering
        \includegraphics[width=\linewidth]{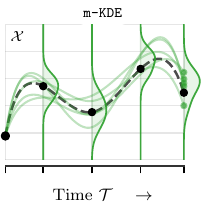}
        \caption{\textit{Viz. of \mkde.}}
        \vspace{-3em}
\end{wrapfigure}
\mkde retains a log-likelihood formulation, which is desirable when the planner benefits from an explicit density over future states.
Rather than scoring each member independently, it constructs a Gaussian-mixture predictive density~\citep{ivanovic2023expanding} that pools all~$S$ samples:
\begin{equation}
    \label{eq:kde_density}
    \hat p_h(\vx^{(n)}_h)
    = \frac{1}{S}\sum_{s=1}^{S}
    \gN\left(\vx^{(n)}_h;\;
    \hat\vx^{(s)}_h,\; h\,\mSigma_\nu\right).
\end{equation}
Each mixture component captures the deterministic prediction error (through its mean~$\hat\vx^{(s)}_h$) and the accumulated aleatoric uncertainty (through the bandwidth~$h\,\mSigma_\nu$), while the mixture itself represents the epistemic uncertainty over~$\mW$.
The loss is the negative marginal log-likelihood of the observed trajectory under the KDE density:
\begin{equation}
    \label{eq:kde_loss}
    \underbracket{
    \ell_{\mathrm{KDE}}\!\big(\{\mW^{(s)}\}_{s=1}^{S},\,
    \bm{\tau}^{(n)}\big)
    \;=\; -\sum_{h=1}^{H}
    \log \hat p_h(\vx^{(n)}_h)}_{\text{Multi-step KDE (\mkde)}}.
\end{equation}

Differentiating through~\eqref{eq:kde_loss}, the log-sum-exp structure assigns each weight sample~$\mW^{(s)}$ a gradient contribution proportional to its \emph{responsibility}---the fraction of the mixture density it explains at the observed state.
Samples whose rollouts pass close to~$\vx^{(n)}_h$ receive large gradients; those that diverge are down-weighted.
This softmax-like reweighting provides an implicit form of trajectory selection that encourages the posterior to concentrate on weight matrices producing accurate multi-step predictions, while the mixture formulation preserves the ability to represent multi-modal predictive distributions that a single Gaussian rollout would collapse.

The log-sum-exp structure of~\eqref{eq:kde_loss} evaluates $-\log\,\E_{q(\mW)}[p(\vx^*_h \mid \mW)]$ rather than $\E_{q(\mW)}[-\log p(\vx^*_h \mid \mW)]$: the logarithm sits \emph{outside} the expectation, coupling all~$S$ weight samples into a single joint evaluation.
No pointwise loss $\ell(\mW)$ exists whose expectation reproduces this quantity, so \rev{\mkde} is \emph{not} an instantiation of the Gibbs VI objective~\eqref{eq:general_obj}.
The correct interpretation is that it applies the log scoring rule directly to the \emph{marginal predictive distribution}
\begin{equation}
    \label{eq:marginal_predictive}
    \bar{p}(\vx_h \mid \vx_0)
    \;=\; \int p(\vx_h \mid \mW, \vx_0)\, q(\mW)\, d\mW,
\end{equation}
estimated via the KDE mixture~\eqref{eq:kde_density}.
By Jensen's inequality, the predictive data-fit is always less pessimistic than a corresponding ELBO term: $\log\E_q[p] \geq \E_q[\log p]$; the mixture receives credit for ensemble diversity in a way that individual-sample scoring cannot but comes with its own computational downsides.
In our experiment, we show that this scoring of the marginal \rev{yields} the best coverage across tasks but can also lead to diverging samples from the tails which we did not observe with losses that optimized the computationally ``easier" Gibbs VI objective.

\paragraph{Relationship between \mkde and \marcrps.}
 
The \marcrps{} shares its autoregressive rollout and linearly growing noise envelope with the KDE objective of~\citet{ivanovic2023expanding}; the two differ in how the resulting ensemble is aggregated and scored.
The \marcrps{} evaluates each weight draw independently: $\ell_{\marcrps}(\mW, \bm{\tau}^{(n)})$ is a pointwise function of a single~$\mW$, and takes the expectation under $q(\mW)$ outside the scoring rule.
This pointwise structure is what makes the \marcrps{} an instantiation of the Gibbs VI objective~\eqref{eq:general_obj}, inheriting the associated PAC-Bayesian interpretation.
The KDE objective instead pools all $S$~weight samples into the Gaussian-mixture predictive $\hat{p}_h = \frac{1}{S}\sum_{s}\gN(\hat{\vx}^{(s)}_h,\, h\,\mSigma_\nu)$ and applies the log scoring rule to this marginal density.
Because the logarithm sits outside the average over ensemble members, the loss couples all draws into a single joint evaluation that falls outside the GVI framework.
The matched \marnll{} disentangles these two differences as it shares the pointwise Gibbs VI structure, rollout, and noise envelope of \marcrps{}, but applies the log score, so comparing \marnll{} with \marcrps{} isolates the scoring rule, while comparing \marnll{} with \mkde{} isolates the aggregation structure.
This structural difference propagates into the gradient: under the \marcrps{}, every weight sample receives an $\mathcal{O}(1)$ gradient signal thanks to the CRPS linear tail behavior, whereas the KDE's log-sum-exp assigns each sample a gradient proportional to its responsibility in the mixture, exponentially down-weighting draws whose rollouts diverge from the observation.
While the CRPS losses under Gibbs VI demonstrated the strongest performance, our results also show very good coverage of \mkde which indicates that a potential extension of \mkde with CRPS might be interesting.

\newpage
\section{Related Work (extended version)}
\label{sec:rw_appendix}

\paragraph{Bayesian neural networks in dynamics learning.}
Probabilistic dynamics models have long been central to model-based control, with Gaussian processes providing strong uncertainty estimates in the low-data regime~\citep{deisenroth2011pilco}.
For neural network dynamics models, probabilistic ensembles (PE)~\citep{lakshminarayanan2017simple} remain popular in model-based reinforcement learning, combining bootstrap ensembles with Gaussian output heads to capture epistemic and aleatoric uncertainty, and underpin a large body of planning and policy-optimization work~\citep{chua2018deep,janner2019trust, frauenknecht2024trust}.
Fully Bayesian alternatives such as MC-dropout~\citep{gal2016dropout} have also been applied to dynamics learning \citep{gal2016improving}, but typically at a higher computational cost for predictive inference and with weaker calibration in practice.
More recently, Bayesian last-layer models~\citep{snoek2015scalable, riquelme2018deep,fiedler2023improved, watson2021latent,harrison2024variational} have emerged as an attractive middle ground: restricting the Bayesian treatment to the final layer preserves tractable inference while still providing useful predictive uncertainty for downstream tasks \citep{harrison2018meta,lew2022safe,davydov2025first,de2026goal}.
Variational Bayesian last layers~\citep{harrison2024variational} instantiate this idea by directly learning a variational posterior of the last layer and have since been applied to Bayesian optimization~\citep{brunzema2025bayesian} and autonomous racing~\citep{brunzema2026vision}.
Our work builds directly on VBLLs but reinterprets them through the lens of Gibbs variational inference~\citep{alquier2016properties, bissiri2016general, knoblauch2022optimization}, decoupling the posterior from the training loss and enabling task-aligned objectives.
Through this, we extend VBLLs to multi-step losses.
BLLs have also been popular in a meta-learning setting--especially for dynamics learning.
Here, the goal is usually to learn the prior distribution $p(\mW)$ and then recursively update it via Kalman filtering \citep{ivanovic2023expanding,levy2025meta} or recursive least squares \cite{harrison2018meta,davydov2025first}.
By design, these already are trained via multi-steps at the cost of back-propagating through the conditioning.
In this work, we do not consider such online adaptation, though it is noteworthy that recursive updates on the learned posteriors are still possible but following the Gibbs VI framework these would have to be gradient-based as there exist in general no closed-form update for an arbitrary loss $\ell$.

\paragraph{Multi-step loss functions.}
The importance of training dynamics models for long-horizon accuracy is well established and some works explicitly demonstrate the downsides of one-step training losses for compounding errors \citep{lambert2022investigating}.
\citet{doerr2017optimizing} argue that multi-step objectives become necessary when the Markov property is violated or when process and observation noise are significant, and 
we further demonstrate that a one-step loss can also exhibit failure modes in very controlled Markov environments.
\citet{lambert2021learning} show that multi-step training improves long-horizon accuracy and sample efficiency in model-based RL.
Closest to our setting, \citet{ivanovic2023expanding} propose a kernel-density multi-step objective with linearly growing noise for trajectory prediction under heteroskedasticity, which we adopt as our \texttt{m-KDE} baseline and which inspired the \marcrps construction.
In contrast to these log-likelihood-based formulations, our CRPS-based losses leverage a strictly proper scoring rule~\citep{matheson1976scoring,gneiting2005calibrated,gneiting2007strictly,grimit2006continuous} that is robust to outliers \cite{selten1998axiomatic}, and are justified within the Gibbs VI framework rather than through a likelihood interpretation.
Such CRPS-based scoring is very popular \citep{berrisch2023crps}, especially in weather forecasting \cite{gneiting2007strictly,lang2026aifs,pacchiardi2024generalized} but often used to directly score marginal distributions \citep{lang2026aifs}.
To the best of our knowledge, we are the first to adopt it to the variational setting through Gibbs VI.
Note that standard CRPS (or in general losses based on standard formulation of the energy distance or MMD) do not fit into the Gibbs VI framework as draws of $\mW$ are not scored independently.

In the context of probabilistic ensembles, \citet{levy2024learning} propose mean propagation for multi-step training over a small horizon ($H=4$) and show improved performance for a structured dynamics model.
Extending multi-step training to probabilistic ensembles in a principled manner would be a natural direction for future work, though the ensemble structure poses additional challenges compared to the Bayesian last-layer setting, as the independence structure across ensemble members differs from the weight posterior in VBLLs making consistent rollouts (for VBLLs by design just a draw of $\mW$) non-trivial.
Here, using approaches as proposed by \citet{frauenknecht2025rollouts} also for model learning can be interesting.
 
For deterministic dynamics models, various works have extended the standard one-step MSE loss to a multi-step setting with mixed results.
\citet{lutter2021learning} demonstrate that while a short multi-step horizon can be beneficial, too long a horizon actually results in worse downstream performance, leading \citet{byravan2021evaluating} to directly build on these short-horizon multi-step models.
To achieve better performance for training with longer horizons, \citet{benechehab2024multi} introduce a weighted sum over time-steps to stabilize gradients and down-weight future predictions; however, also with mixed results.
We find that for (Gibbs) VBLLs, we did not need to specify additional weighting which would result from tempered likelihoods, as the VBLL setup seems to provide sufficient flexibility, especially in combination with the CRPS losses, accommodate and then also optimize for horizons of up to 100 time steps (see Appendix~\ref{sec:app-additional-results}).
For Gaussian processes, moment matching and uncertainty propagation via linearization have been used to train with multi-step objectives~\citep{deisenroth2011pilco,doerr2017optimizing}.
In that setting, the set of learnable parameters is significantly smaller; in the regime of overparameterized neural networks, it is even more important to provide informative multi-step training signal to learn good and reliable features.
 
\paragraph{Objective mismatch and decision-aware model learning.}
A fundamental tension in model-based control is that minimizing prediction error does not necessarily maximize downstream task performance.
\citet{lambert2020objective} coin the term \emph{objective mismatch} and demonstrate empirically that model accuracy and control performance can be poorly correlated, arguing that the model learning objective should be informed by the decision-making pipeline it serves.
This observation has motivated a growing line of work on decision-aware model learning, recently surveyed by \citet{wei2023unified}.
\citet{farahmand2017value} introduce value-aware model learning, which replaces the standard maximum-likelihood objective with a loss that incorporates the structure of the value function space, ensuring that the model is accurate where it matters for policy evaluation.
This idea was further developed in~\citet{farahmand2018iterative}, which proposes an iterative scheme that alternates between value estimation and model refinement.
\citet{d2020gradient} take a complementary perspective: their gradient-aware model-based policy search algorithm re-weights the model-learning objective according to policy-gradient norms, focusing model accuracy on state-action regions where the policy gradient is large.
Similarly, \citet{abachi2020policy} derives policy-aware model learning, which aligns the model-learning loss with the structure of the policy gradient estimator.
More recently, \citet{voelcker2022value} propose value-gradient weighted model learning (VaGraM), which rescales the prediction loss with gradient information from the current value function estimate and provides a thorough empirical comparison of decision-aware approaches.
In the vision-based setting, \citet{nair2020goal} propose goal-aware prediction learning, which conditions the dynamics model on the goal state, directing model capacity toward task-relevant aspects of the observation.

These approaches mostly operate in the deterministic setting with point-estimate models and focus on coupling model learning directly with policy optimization.
Our work shares the underlying motivation---the training loss should reflect how the model is used---but takes a different and complementary route.
Rather than incorporating a value function or policy gradient into the model-learning objective, we remain in the probabilistic prediction setting and use the Gibbs VI framework to replace the likelihood with losses that are structurally aligned with the deployment regime of the model (multi-step rollouts, robust scoring).
Notably, the two perspectives are not mutually exclusive: because the Gibbs VI objective~\eqref{eq:gibbs} accepts any pointwise loss $\ell(\mW, \tau^{(n)})$, decision-aware reweightings such as VaGraM can be incorporated by simply scaling each per-step CRPS term by the local value-gradient magnitude, yielding a value-aware multi-step scoring rule that retains a PAC-Bayesian interpretation.
Exploring such combinations in a closed-loop control setting is a promising direction for future work.

\newpage
\section{Proof of Proposition~\ref{prop:linear_exactness_rewrite}}
\label{sec:proof_prop}

For convenience, we restate the proposition in a slightly expanded form that makes the leading correction term explicit; the bound in the main text follows as an immediate corollary.

\begin{proposition*}
\label{prop:linear_exactness_rewrite_refined}
Suppose $\phi_\vtheta(\vx) = \mPhi \vx + \vb$ and define $\mA \defeq \mI + \mW^\top \mPhi$.
Then for all $h \ge 1$ the predictive distribution is exactly Gaussian,
\begin{equation}
    p(\vx_h \mid \mW, \vx_0) = \gN\!\big(\hat{\vx}_h(\mW),\, \mC_h\big),
    \qquad
    \mC_h \defeq \sum_{k=0}^{h-1} \mA^k \mSigma_\nu (\mA^k)^\top.
\end{equation}
The covariance collapses to $\mC_h = h\,\mSigma_\nu$ whenever $\mA\,\mSigma_\nu\,\mA^\top = \mSigma_\nu$ (\eg $\mA$ orthogonal and $\mSigma_\nu = \sigma^2 \mI$). More generally, writing $\mA = \mI + \mE$ with $\mE = \mW^\top \mPhi$,
\begin{equation}
    \label{eq:Ch_expansion}
    \mC_h
    = h\,\mSigma_\nu
    + \frac{h(h-1)}{2}\big(\mE\,\mSigma_\nu + \mSigma_\nu\,\mE^\top\big)
    + \mathcal{O}\!\big(h^3 \|\mE\|^2 \|\mSigma_\nu\|\big),
\end{equation}
valid whenever $h\,\|\mE\|$ is small.
\end{proposition*}

\begin{proof}
Under affine features, the stochastic dynamics reduce to
$
    \vx_{h+1} = \mA\,\vx_h + \mW^\top\vb + \bm{\nu}_h,
$
which is a linear time-invariant system driven by i.i.d.\ Gaussian noise $\bm{\nu}_h \sim \mathcal{N}(\bm{0}, \mSigma_\nu)$. Unrolling from $\vx_0$,
\begin{equation}
    \label{eq:unrolled}
    \vx_h = \mA^h \vx_0 + \sum_{k=0}^{h-1}\mA^k\,\mW^\top\vb
    + \sum_{k=0}^{h-1}\mA^{h-1-k}\,\bm{\nu}_k.
\end{equation}

\emph{Part (i).}\;
The deterministic rollout satisfies
$\hat{\vx}_h(\mW) = \mA^h \vx_0 + \sum_{k=0}^{h-1}\mA^k\,\mW^\top\vb$.
Since $\E[\bm{\nu}_k]=\bm{0}$, taking expectations in \eqref{eq:unrolled} yields
$\E[\vx_h \mid \mW, \vx_0] = \hat{\vx}_h(\mW)$.

\emph{Part (ii).}\;
Subtracting the mean from \eqref{eq:unrolled} gives
\begin{equation}
    \vx_h - \E[\vx_h \mid \mW, \vx_0]
    = \sum_{k=0}^{h-1}\mA^{h-1-k}\,\bm{\nu}_k.
\end{equation}
Since the $\bm{\nu}_k$ are independent with covariance $\mSigma_\nu$, the conditional covariance is
\begin{equation}
    \mC_h
    = \sum_{k=0}^{h-1}\mA^{h-1-k}\,\mSigma_\nu\,(\mA^{h-1-k})^\top
    = \sum_{k=0}^{h-1}\mA^k\,\mSigma_\nu\,(\mA^k)^\top,
\end{equation}
where the second equality follows by reindexing $k \mapsto h-1-k$.

\emph{Exactness.}\;
If $\mA\,\mSigma_\nu\,\mA^\top = \mSigma_\nu$, then by induction $\mA^k\,\mSigma_\nu\,(\mA^k)^\top = \mSigma_\nu$ for all $k \ge 0$, and therefore $\mC_h = h\,\mSigma_\nu$ exactly. This holds in particular when $\mA$ is orthogonal and $\mSigma_\nu = \sigma^2 \mI$, since then $\mA^k (\mA^k)^\top = \mI$.

\emph{Expansion.}\;
Write $\mA = \mI + \mE$. The binomial expansion gives
\begin{equation}
    \mA^k = (\mI + \mE)^k = \mI + k\,\mE + \mR_k,
    \qquad
    \|\mR_k\| = \mathcal{O}\!\big(k^2 \|\mE\|^2\big).
\end{equation}
Substituting into each summand and keeping terms up to first order in $\mE$,
\begin{align}
    \mA^k \mSigma_\nu (\mA^k)^\top
    &= (\mI + k\mE + \mR_k)\,\mSigma_\nu\,(\mI + k\mE + \mR_k)^\top \\
    &= \mSigma_\nu
    + k\big(\mE\,\mSigma_\nu + \mSigma_\nu\,\mE^\top\big)
    + \mathcal{O}\!\big(k^2 \|\mE\|^2 \|\mSigma_\nu\|\big).
\end{align}
Summing over $k = 0, \dots, h-1$ and using
$\sum_{k=0}^{h-1} 1 = h$, $\sum_{k=0}^{h-1} k = \tfrac{h(h-1)}{2}$, and
$\sum_{k=0}^{h-1} k^2 = \mathcal{O}(h^3)$,
\begin{equation}
    \mC_h
    = h\,\mSigma_\nu
    + \frac{h(h-1)}{2}\big(\mE\,\mSigma_\nu + \mSigma_\nu\,\mE^\top\big)
    + \mathcal{O}\!\big(h^3 \|\mE\|^2 \|\mSigma_\nu\|\big),
\end{equation}
which establishes \eqref{eq:Ch_expansion}.
\end{proof}

\begin{remark}
\label{rem:main_text_bound}
The bound stated in the main text follows immediately from \eqref{eq:Ch_expansion}: when $h\,\|\mE\|$ is small, the $\mathcal{O}(h^3\|\mE\|^2\|\mSigma_\nu\|)$ term is dominated by the leading correction, and therefore
\begin{equation}
    \|\mC_h - h\,\mSigma_\nu\| = \mathcal{O}\!\big(h^2\,\|\mE\|\,\|\mSigma_\nu\|\big).
\end{equation}
In the residual parameterization $\mA = \mI + \mW^\top\mPhi$ with small time step $\Delta t$, one has $\|\mE\| = \mathcal{O}(\Delta t)$, so the condition $h\,\|\mE\|$ small corresponds to a bounded rollout time $h\,\Delta t$.
\end{remark}

\newpage
\section{Closed-Form CRPS for Gaussian Predictives}
\label{app:crps_derivation}

The closed-form expression of the CRPS for Gaussian predictive distributions is well known; see, e.g., \citep{gneiting2005calibrated,gneiting2007strictly}.
We briefly include the derivation here for completeness.

Starting from the energy form,
\begin{equation}
    \label{eq:crps_energy_app}
    \operatorname{CRPS}(P, y)
    \;=\;
    \E_P|X - y|
    \;-\;
    \tfrac{1}{2}\,\E_P|X - X'|,
\end{equation}
with $X, X' \overset{\text{iid}}{\sim} P$, we evaluate both terms for $P = \gN(\mu, \sigma^2)$.

\paragraph{Term 1: $\E|X - y|$.}

Let $Z \sim \gN(0,1)$ and write $X = \mu + \sigma Z$, with $z \defeq (y - \mu)/\sigma$. Then
\[
\E|X - y| = \sigma\,\E|Z - z|.
\]
Using standard manipulations,
\begin{align*}
\E|Z - z|
&= \int_{-\infty}^{z} (z - t)\varphi(t)\,dt
  + \int_{z}^{\infty} (t - z)\varphi(t)\,dt \\
&= z\Phi(z) + \varphi(z)
   + \varphi(z) - z\big(1 - \Phi(z)\big) \\
&= 2\varphi(z) + z\big(2\Phi(z) - 1\big),
\end{align*}
where we used $\int t\varphi(t)\,dt = -\varphi(t)$. Hence,
$
    \E|X - y|
    = \sigma\Big[2\varphi(z) + z\big(2\Phi(z) - 1\big)\Big].
$

\paragraph{Term 2: $\E|X - X'|$.}

Since $X - X' \sim \gN(0, 2\sigma^2)$, we write $X - X' = \sigma\sqrt{2}\,Z$ with $Z \sim \gN(0,1)$, giving
\[
\E|X - X'|
= \sigma\sqrt{2}\,\E|Z|
= \sigma\sqrt{2}\cdot \sqrt{\frac{2}{\pi}}
= \frac{2\sigma}{\sqrt{\pi}}.
\]

Substituting both derived terms into~\eqref{eq:crps_energy_app} yields the closed form from \eqref{eq:crps_gaussian}.

\newpage
\section{Additional Results}
\label{sec:app-additional-results}

\subsection{Extended Statistical evaluation of the chaotic systems benchmark}
\label{sec:app-statistical-eval}

To complement the average rank summary in Figure~\ref{fig:results_summary}, we provide a formal statistical comparison across all 198 (environment, metric) evaluation blocks from Section~\ref{sec:results-benchmark}. First we conducted a Friedman test that rejects the null hypothesis that all methods perform equally ($\chi^2 = 314.1$, $p < 10^{-65}$) and then followed this with post-hoc comparisons.
Table~\ref{tab:rank-distribution} reports the rank distribution and these pairwise post-hoc comparisons.

All pairwise differences are significant at $\alpha = 0.05$ under the Holm-corrected Wilcoxon signed-rank test; under the more conservative Nemenyi test, only the differences between \mkde{} and \selbo{} as well as between \mkde{} and \marcrps{} are not significant.
In particular, both CRPS-based losses significantly outperform their likelihood-based counterparts (\mfcrps{} vs.\ \melbo{} and \marcrps{} vs.\ \marnll{}).
For the autoregressive pair, this advantage stems from training stability.
\marnll{} diverged on all 13 chaotic systems and is therefore ranked last in 143 out of 198 cases, whereas \marcrps{} trained stably throughout; on the five systems where \marnll{} converged, the two are statistically indistinguishable.
Notably, among the stably trained losses, \selbo{} and \mkde{} have the worst mean ranks, underscoring the limitations of single-step training and of the marginal mixture objective, respectively.

\begin{table}[h]
\centering
\caption{\textit{Statistical evaluation across 198 (environment, metric) blocks.} \textit{Left:} Rank distribution (lower mean rank is better). \textit{Right:} Post-hoc pairwise comparisons (Nemenyi test, critical difference $\Delta r = 0.536$, and Holm-corrected Wilcoxon signed-rank test); pairs below the line are not significant at $\alpha = 0.05$ under the Nemenyi test. Friedman test: $\chi^2 = 314.1$, $p < 10^{-65}$. Diverged runs (143 blocks of \texttt{m-arNLL}) are ranked last.}
\label{tab:rank-distribution}
\resizebox{\textwidth}{!}{%
\begin{tabular}{lcccrrrrrr}
\toprule
Method & Mean & Med. & Std & 1st & 2nd & 3rd & 4th & 5th & 6th \\
\midrule
\texttt{m-CRPS} & 2.12 & 2 & 1.26 & 85 & 49 & 35 & 16 & 11 & 2 \\
\texttt{m-NLL} & 2.83 & 2 & 1.48 & 46 & 56 & 23 & 33 & 39 & 1 \\
\texttt{m-arCRPS} & 3.24 & 3 & 1.51 & 28 & 50 & 27 & 47 & 32 & 14 \\
\texttt{s-NLL} & 3.78 & 4 & 1.32 & 9 & 17 & 66 & 50 & 29 & 27 \\
\texttt{m-KDE} & 3.83 & 4 & 1.43 & 20 & 20 & 31 & 40 & 77 & 10 \\
\texttt{m-arNLL} & 5.21 & 6 & 1.47 & 10 & 6 & 16 & 12 & 10 & 144 \\
\bottomrule
\end{tabular}%
\qquad
\begin{tabular}{llrcc}
\toprule
Method A & Method B & $\Delta\bar{r}$ & $p_{\text{Nemenyi}}$ & $p_{\text{Holm}}$ \\
\midrule
\texttt{m-arNLL} & \texttt{m-CRPS} & 3.10 & ${<}\,10^{-10}$ & ${<}\,10^{-10}$ \\
\texttt{m-arNLL} & \texttt{m-NLL} & 2.38 & ${<}\,10^{-10}$ & ${<}\,10^{-10}$ \\
\texttt{m-arNLL} & \texttt{m-arCRPS} & 1.97 & ${<}\,10^{-10}$ & ${<}\,10^{-10}$ \\
\texttt{m-KDE} & \texttt{m-CRPS} & 1.71 & ${<}\,10^{-10}$ & ${<}\,10^{-10}$ \\
\texttt{s-NLL} & \texttt{m-CRPS} & 1.66 & ${<}\,10^{-10}$ & ${<}\,10^{-10}$ \\
\texttt{m-arNLL} & \texttt{s-NLL} & 1.43 & ${<}\,10^{-10}$ & ${<}\,10^{-10}$ \\
\texttt{m-arNLL} & \texttt{m-KDE} & 1.38 & ${<}\,10^{-10}$ & ${<}\,10^{-10}$ \\
\texttt{m-arCRPS} & \texttt{m-CRPS} & 1.12 & ${<}\,10^{-7}$ & ${<}\,10^{-9}$ \\
\texttt{m-KDE} & \texttt{m-NLL} & 1.00 & ${<}\,10^{-5}$ & ${<}\,10^{-5}$ \\
\texttt{s-NLL} & \texttt{m-NLL} & 0.95 & ${<}\,10^{-5}$ & ${<}\,10^{-6}$ \\
\texttt{m-NLL} & \texttt{m-CRPS} & 0.71 & $2 \times 10^{-3}$ & ${<}\,10^{-5}$ \\
\texttt{m-KDE} & \texttt{m-arCRPS} & 0.59 & $0.021$ & $8 \times 10^{-4}$ \\
\texttt{s-NLL} & \texttt{m-arCRPS} & 0.54 & $0.047$ & $3 \times 10^{-3}$ \\
\midrule
\texttt{m-arCRPS} & \texttt{m-NLL} & 0.41 & $0.249$ & $0.073$ \\
\texttt{m-KDE} & \texttt{s-NLL} & 0.05 & $1.000$ & $0.670$ \\
\bottomrule
\end{tabular}%
}
\end{table}

\subsection{Intuition on the different learned feature representations}

\begin{figure}
    \centering
    \includegraphics[width=\linewidth]{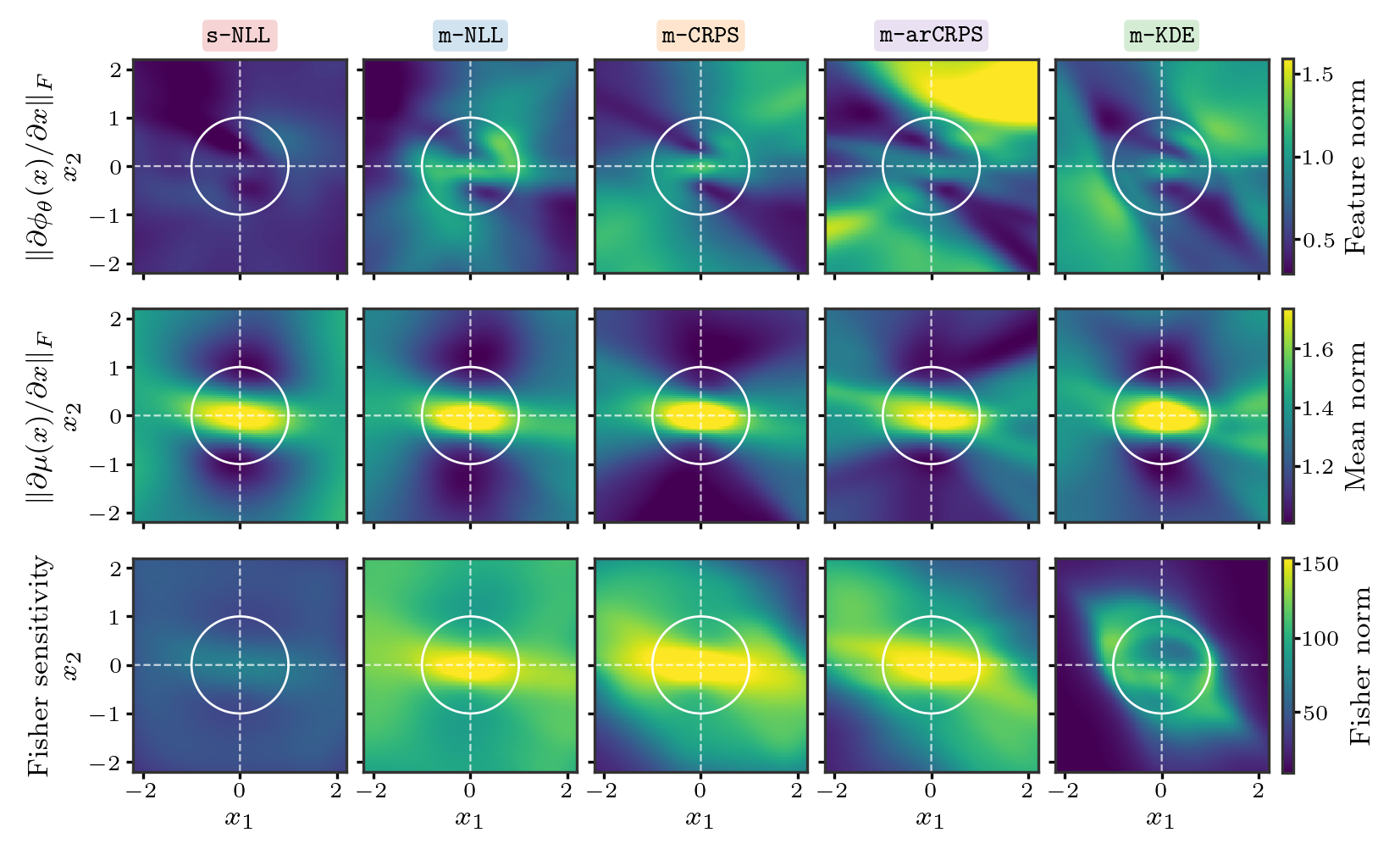}%
    \vspace{-1em}
    \caption{\textit{Feature sensitivity (top), mean prediction sensitivity (middle), and Fisher sensitivity (bottom) for the learned model on the VSLC (Fig.~\ref{fig:vslc}).}
    All models are evaluated on the same data yet there are distinct differences in behavior, most prominently in feature and Fisher sensitivity.}
    
    \label{fig:features}
\end{figure}

Figure~\ref{fig:features} visualizes how different training losses shape the learned representation $\phi_\theta(x)$ and its induced sensitivities across the state space of the VSLC.
A key property of the VSLC is its state-dependent angular velocity, which induces spatial heterogeneity: the upper-right and lower-left regions correspond to fast zones (also see Fig.~\ref{fig:vslc}), while the remaining regions are slow. Due to the inverse relationship between data density and velocity, the training data is heavily biased toward the slow regions.

\paragraph{Feature sensitivity.}
The top row reveals a strong difference across losses.
\selbo shows a low feature sensitivity which slightly larger sensitivity in the fast zones compared to the slow zones.
In contrast, multi-step objectives show more diverse feature fields.
Especially the CRPS losses and \mkde demonstrate increased sensitivity in the fast zones (upper-right and lower-left), where accurate modeling is most critical for long-horizon predictions.
It indicates that the multi-step losses encourage the representation to allocate capacity to dynamically important but under-sampled regions.

\paragraph{Mean prediction sensitivity.}
The middle row shows that all methods learn a similar structure for the mean dynamics, dominated by a strong horizontal band around $x_2 \approx 0$.
This suggests that, despite differences in representation, the mean predictor is largely constrained by the residual model structure as well as the defined prior.
There are however still clear observable differences between the different losses.

\paragraph{Fisher sensitivity.}
The bottom row highlights how uncertainty interacts with the learned features.
We define the Fisher sensitivity as
\begin{equation}
    \label{eq:fisher_sensitivity}
    \mathcal{S}_{\mathrm{F}}(\vx)
    \;\defeq\;
    \left(
    \sum_{j=1}^{d_x}
    \frac{\big\|\nabla_{\vx}\mu_j(\vx)\big\|_2^2}{\sigma_j^2(\vx)}
    +
    2\sum_{j=1}^{d_x}
    \big\|\nabla_{\vx}\log\sigma_j(\vx)\big\|_2^2
    \right)^{1/2},
\end{equation}
The Fisher sensitivity differs substantially across losses, reflecting how the posterior over the last layer couples with the representation.
While \selbo yields relatively uniform and low Fisher sensitivity due to the high noise estimates, multi-step and CRPS-based losses produce structured patterns.
In particular, \mfcrps exhibits a high Fisher sensitivity at the transitions from slow to fast dynamic sections of the state space while \mkde shows a very pronounced sensitivity around the limit cycle.
\label{sec:different_features}

In the end, the important evaluation is the one from Figure~\ref{fig:vslc} but still these results show that using the same data, the loss clearly influences both the learned features as well as the learned variational posterior.

\begin{figure}[t]
    \begin{subfigure}[b]{\linewidth}
        \centering
        \includegraphics[width=\linewidth]{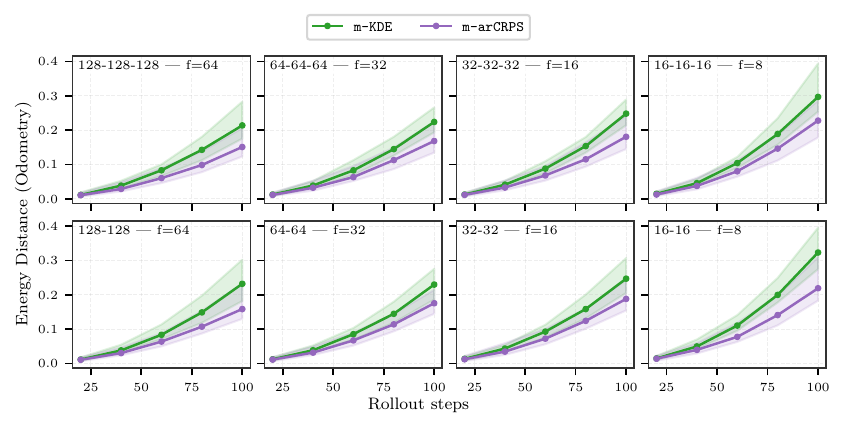}%
        \vspace{-8pt}
        \caption{\textit{Performance of \mkde and \marcrps under varying model size.}
        We evaluate both losses on different sizes of the backbone. \marcrps consistently outperforms \mkde across the full horizon of 100 time steps.}
        \label{fig:model_capacity_wheelbot}
    \end{subfigure}
    
    \centering    
    \begin{subfigure}[t]{0.24\linewidth}
        \centering
        \includegraphics[width=1\linewidth]{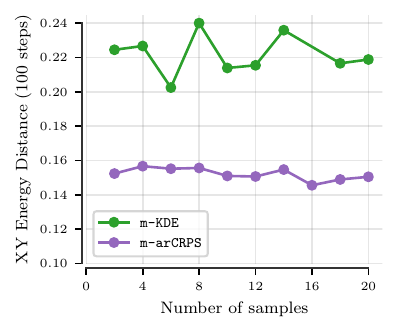}%
        \caption{Performance.}
    \end{subfigure}
    \hfill
    \begin{subfigure}[t]{0.24\linewidth}
        \centering
        \includegraphics[width=1\linewidth]{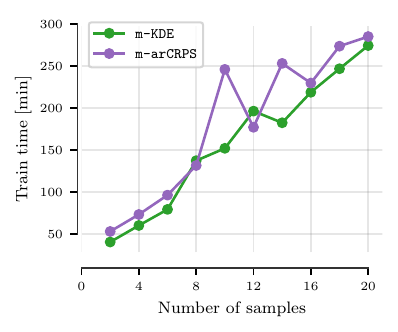}
        \caption{Training time.}
    \end{subfigure}
    \hfill
    \begin{subfigure}[t]{0.24\linewidth}
        \centering
        \includegraphics[width=1.05\linewidth]{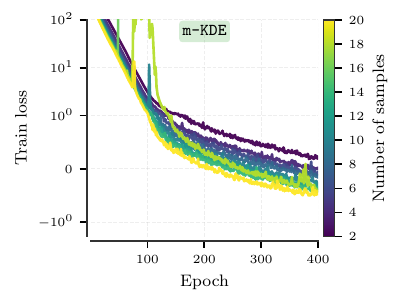}
        \caption{Training losses.}
    \end{subfigure}
    \hfill
    \begin{subfigure}[t]{0.24\linewidth}
        \centering
        \includegraphics[width=1.05\linewidth]{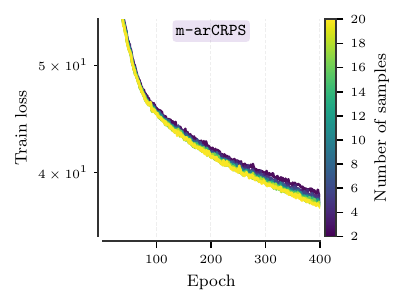}
        \caption{Training losses.}
    \end{subfigure}%
    \caption{\textit{Ablation of model size (top) and the sample size $S$ (bottom) on Mini Wheelbot data.} }%
    \label{fig:wb_ablation}
\end{figure}%

\subsection{Sensitivity of \mkde and \marcrps to Model Capacity}
\label{sec:sensitivity_model_capa}

In all experiments, we fixed the model size and capacity and did not perform any dedicated tuning of these aspects.
Nevertheless, studying how model size and capacity influence predictive performance is both relevant and insightful.
In particular, it is plausible that certain loss functions are better able to exploit increased model capacity.
To investigate this, we conduct an ablation study on model capacity for both \mkde{} and \marcrps{} using the Mini Wheelbot.
The results are summarized in Fig.~\ref{fig:wb_ablation}.

For both three- and two-layer backbone (top and bottom row, respectively), we observe that increasing model capacity consistently improves performance, measured here by the energy distance on the $xy$-plane across multiple rollouts.
Across all model sizes, \marcrps{} outperforms \mkde{}.
Interestingly, the smallest model trained with \marcrps{} (bottom right) achieves performance comparable to the largest model trained with \mkde{} (top left).
While the absolute model sizes are tiny by modern standards, a reduction of up to 70$\times$ in the number of parameters is highly significant, particularly for deployment on embedded hardware.

\subsection{Sensitivity of \mkde and \marcrps to Sample Size \texorpdfstring{$S$}{S}}

In all of our experiments, we use $S=10$ and did not further tune this for any of the methods.
In Fig.~\ref{fig:wb_ablation}~(b)--(e), we ablate the sensitivity of the loss functions to different sample sizes $S$ on the Mini Wheelbot data.
We can see that both models perform comparable across different $S$.
In Fig.~\ref{fig:wb_ablation}~(c) we can further see the $\gO(S)$ scaling of the loss functions where small perturbations from the linear trend are due to minor cluster anomalies.
Lastly, we can observe that the training with \mkde becomes a bit more unstable for larger $S$ actually resulting in a diverged training for $S=16$.
Training with \marcrps is very stable and behaves similar for the different $S$.

\subsection{Sensitivity of \mkde and \marcrps to Horizon \texorpdfstring{$H$}{H}}

\begin{figure}
    \centering
    \includegraphics[width=0.49\linewidth]{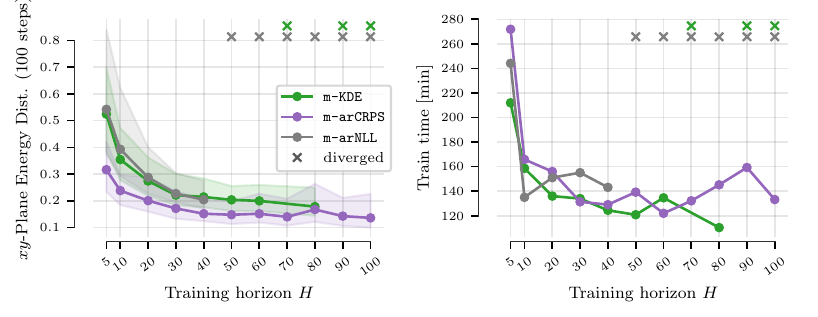}
    \includegraphics[width=0.49\linewidth]{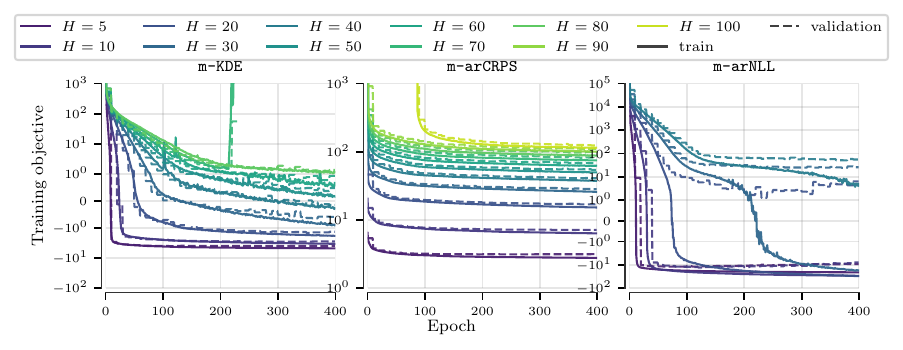}
    \caption{\textit{Sensitivity to different horizon $H$.} A longer horizon during training increases performance. Training time for small $H$ increases as the effective data set size increase.}
    \label{fig:sweep_over_h}
\end{figure}

In Fig.~\ref{fig:sweep_over_h}, we compare the performance of \mkde and \marcrps on the Mini Wheelbot data for different horizons $H$.
We find that generally increasing the horizon shows improved performance but it plateaus for $H>40$.
Furthermore, we find that for large horizon, training with \mkde was numerically more unstable resulting in the training runs of $H\in\{70,90,100\}$ to diverge.
The matched \marnll trains up to $H=40$ but diverges for all larger horizons, while \marcrps remains stable across all tested horizons.
In general, we observed that training with a too long horizon can be challenging but having a robust scoring and with this bounded gradients on the noise helped in such long-horizon training scenarios.

\subsection{Effect of the regularization factor for \mfcrps and \marcrps}
\label{sec:reg_weight}

\begin{figure}
    \centering
    \includegraphics[width=\linewidth]{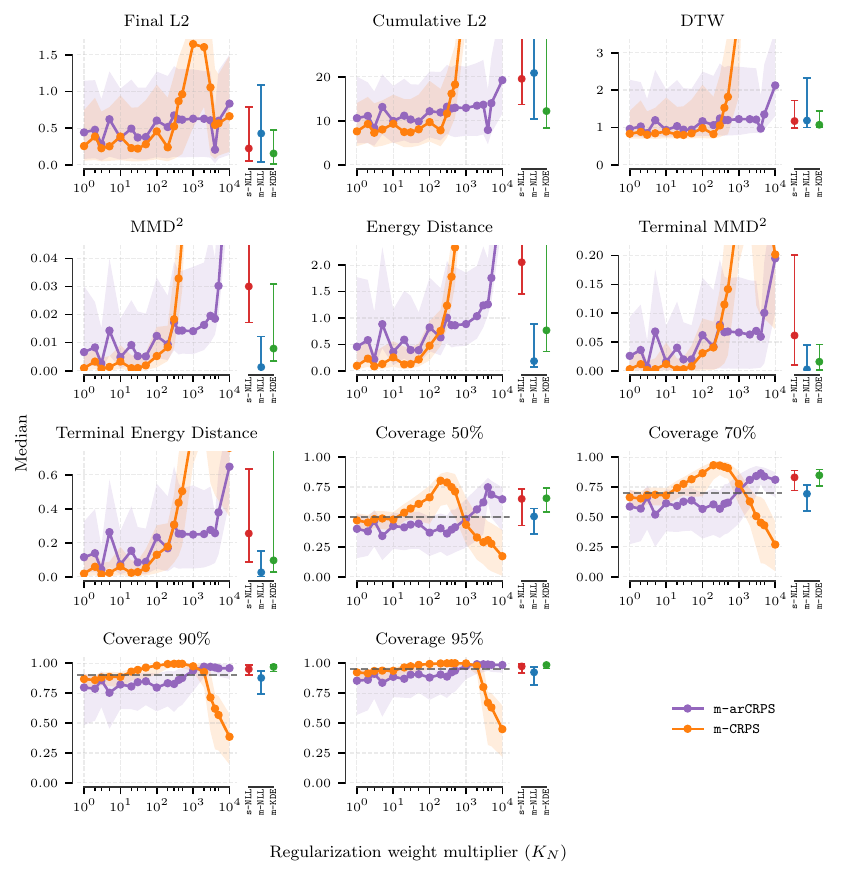}%
    \vspace{-1em}
    \caption{\textit{Influence of changing the KL weight $\lambda$ for the CRPS baselines.}}
    \vspace{-1em}
    \label{fig:kl}
\end{figure}

All the results in the main paper were generated with $\lambda = 1 /N$.
However, we observe that the autoregressive \marcrps can be overconfident in terms of coverage (also see Tab.~\ref{tab:eval_new_v1_test_booktabs_part1}--\ref{tab:eval_new_v1_test_booktabs_part4}).
One way to increase coverage that directly aligns with the Gibbs VI framework is to increase the regularization by a factor $K_N$ resulting in $\lambda = K_N / N$ for the different methods.
Note that also \citep{brunzema2026vision} use $K_N=10$ for the one-step ELBO.
In Fig.~\ref{fig:kl}, we show how increasing $K_N$ influences coverage as well as performance on the VSLC environment, where \marcrps tended to be overconfident.
As the scale of the losses differ between \mfcrps and \marcrps, the magnitudes in which $K_N$ needs to be adapted is different for the two methods.
We find that for \mfcrps, setting $K_N=10$ gives good coverage and still strong performance while for \marcrps $K_N=10^3$ demonstrates strong coverage at the cost however of slightly reduced performance.

\clearpage
\subsection{Gradient norms during optimization}
\label{sec:grad_norms}

\begin{figure}[t]
    \centering
    \includegraphics[width=\linewidth]{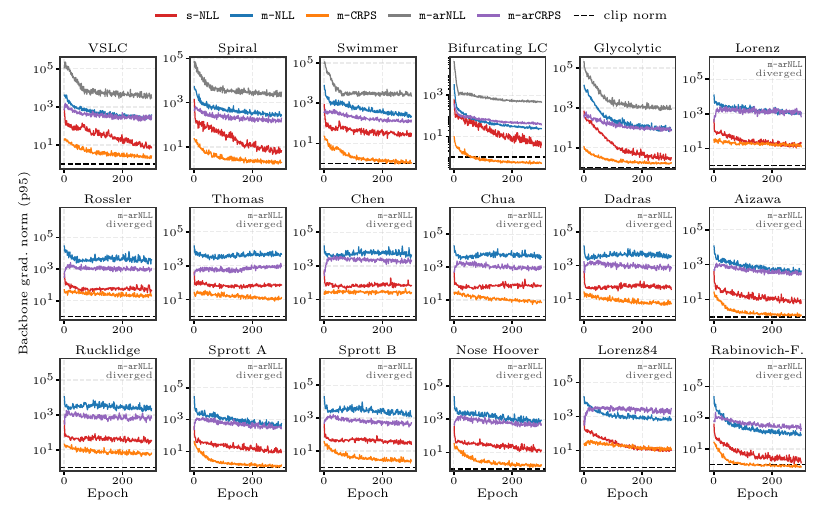}
    \caption{\textit{Backbone gradient norms during training.}
    95th percentile of the pre-clipping backbone gradient norm per epoch for all objectives across all benchmark environments; the dashed line marks the clipping threshold.
    \marnll develops non-finite gradients on all chaotic systems and is therefore only shown for the non-chaotic environments.
    CRPS attenuates but does not prevent the gradient growth induced by autoregressive training.}
    \label{fig:grad_norms}
\end{figure}

The bounded derivatives of the CRPS in Sec.~\ref{sec:likelihood_objectives} hold at the score level, \ie, with respect to the predictive location and scale.
They do not bound gradients through the backbone and, under autoregressive evaluation, backpropagation through time additionally multiplies rollout Jacobians.
To assess how the scoring rule affects optimization in practice, we log the pre-clipping gradient norm of the backbone parameters at every training step for all objectives on all benchmark environments and report its 95th percentile per epoch in Fig.~\ref{fig:grad_norms}.
All objectives use the same global gradient clipping threshold, which we adopt from the original VBLL implementation~\citep{harrison2024variational} and did not tune for any objective (\cf Appendix~\ref{sec:details} for all the implementation details).

\paragraph{Teacher forcing.}
Under teacher forcing, the scoring rule alone already has a strong effect on the gradient magnitude: \mfcrps has by far the smallest backbone gradients, whereas \melbo yields gradients that are orders of magnitude larger.
Since the per-step contributions only add up without being propagated through a rollout, these large gradients remain benign and both objectives train successfully on all environments.

\paragraph{Autoregressive training.}
Once the predictions are fed back along the rollout, gradients compound through time.
Consequently, \marcrps exhibits substantially larger backbone gradients than its teacher-forced counterpart \mfcrps.
Here CRPS attenuates gradient growth under backpropagation through time.
Nevertheless, the gradients of \marcrps remain of the same order as, and typically below, those of the teacher-forced \melbo, and training remains finite on all environments.
In contrast, the matched \marnll, which differs from \marcrps only in the scoring rule, exhibits the largest gradients on all non-chaotic systems and develops non-finite gradients on all chaotic systems.
Since all objectives share the same clipping threshold, clipping alone is insufficient to prevent the instability of the autoregressive log score.
We observe the same pattern on the Mini Wheelbot horizon sweep (Fig.~\ref{fig:sweep_over_h}), where \marnll trains at shorter horizons but diverges at longer ones, while \marcrps remains stable across all tested horizons.

\begin{figure}[p]
    \centering
    \includegraphics[width=\linewidth]{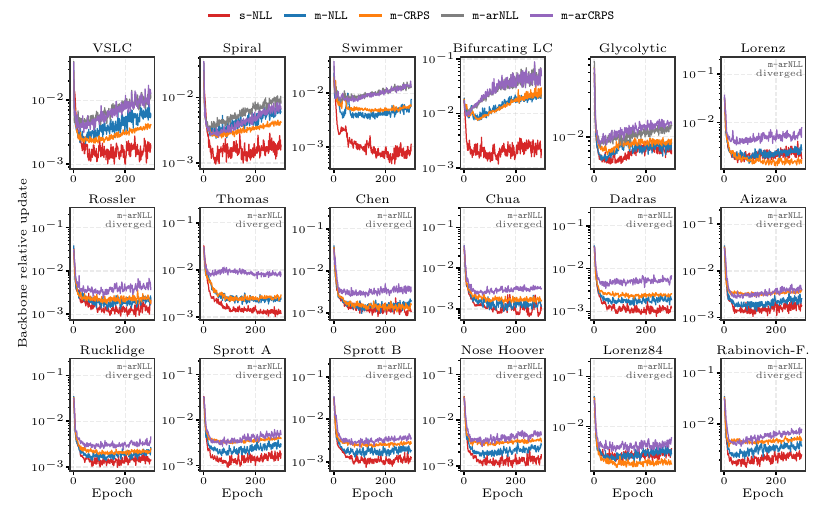}
    \caption{\textit{Relative update of the backbone parameters per epoch.}
    \marnll develops non-finite gradients on all chaotic systems and is therefore only shown for the non-chaotic environments.}
    \label{fig:rel_updates_backbone}
\end{figure}

\begin{figure}[p]
    \centering
    \includegraphics[width=\linewidth]{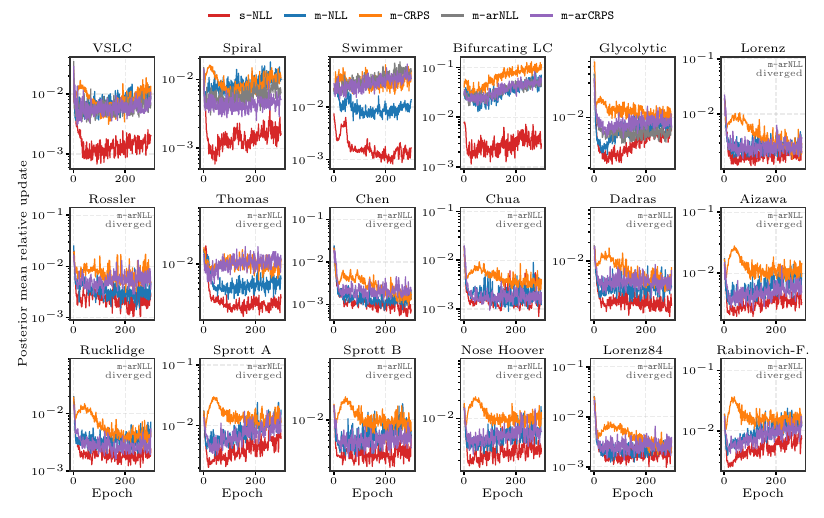}
    \caption{\textit{Relative update of the posterior mean of the last-layer weights per epoch.}
    The posterior mean keeps updating on a similar relative scale as the backbone (\cf Fig.~\ref{fig:rel_updates_backbone}).}
    \label{fig:rel_updates_mean}
\end{figure}

\begin{figure}[p]
    \centering
    \includegraphics[width=\linewidth]{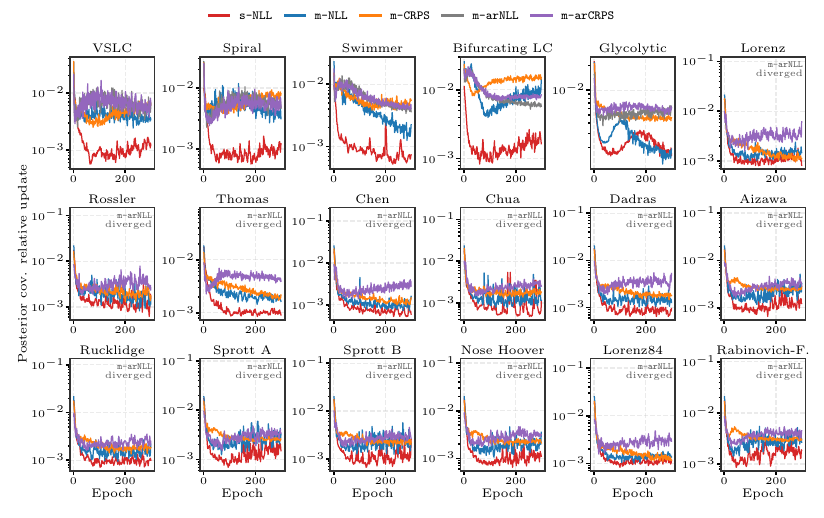}
    \caption{\textit{Relative update of the posterior covariance of the last-layer weights per epoch.}
    As for the posterior mean, the covariance does not stall while the backbone continues to update (\cf Fig.~\ref{fig:rel_updates_backbone}).}
    \label{fig:rel_updates_cov}
\end{figure}

\paragraph{Joint optimization of backbone and posterior.}
Since the backbone and the variational posterior are trained jointly, the last layer performs inference on a representation that shifts during training.
To check whether the posterior keeps up with this representation drift, we track the relative change of the backbone parameters, the posterior mean, and the posterior covariance between consecutive epochs, \ie, the norm of the update divided by the norm of the respective parameters (Figs.~\ref{fig:rel_updates_backbone}--\ref{fig:rel_updates_cov}).
Across all objectives and environments, the posterior mean and covariance continue to update on a similar relative scale as the backbone throughout training.
In particular, we do not observe the posterior stalling while the representation continues to move, \ie, the posterior does not decouple from the evolving features.
We note that, since AdamW normalizes the update by a running estimate of the gradient scale and we use a constant learning rate, these quantities are largely insensitive to the gradient magnitudes in Fig.~\ref{fig:grad_norms} and are not a measure of convergence.

\clearpage
\begin{figure}
    \centering
    \includegraphics[width=\linewidth]{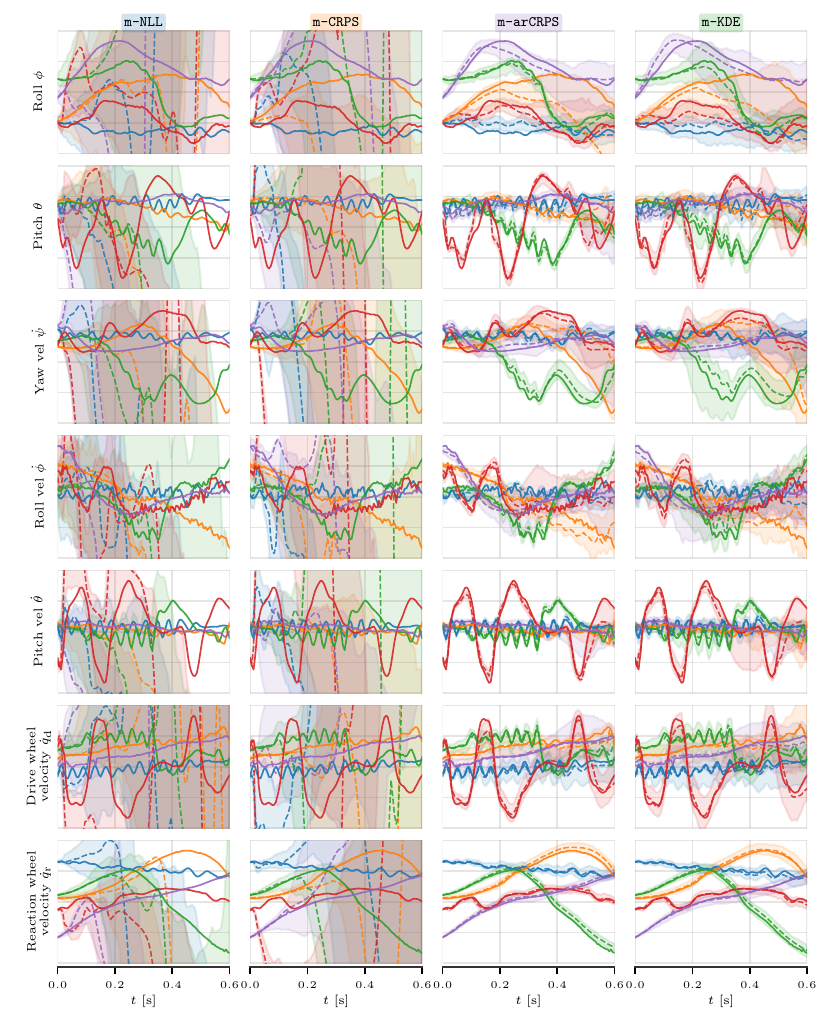}
    \caption{\textit{Rollouts on Mini Wheelbot data.} Figure shows the complete set of states to complement the rollouts shown in Fig.~\ref{fig:wb_results} in the main body of the paper.}
    \label{fig:full_state_rollout}
\end{figure}

\clearpage
\label{sec:full_results_chaotic}

\begin{table*}[t]
\centering
\caption{\textit{Results on the benchmarking of Sec.~\ref{sec:results-benchmark}.} (Part 1/4)}
\label{tab:eval_new_v1_test_booktabs_part1}
\sisetup{group-separator={,},group-minimum-digits=4,detect-weight=true,detect-inline-weight=math}
\footnotesize
\setlength{\tabcolsep}{4pt}
\resizebox{\textwidth}{!}{%
\begin{tabular}{clcccccc}
\toprule
Environment & Metric & \texttt{s-NLL} & \texttt{m-NLL} & \texttt{m-CRPS} & \texttt{m-arNLL} & \texttt{m-arCRPS} & \texttt{m-KDE} \\
\midrule
\multirow[c]{13}{*}{\rotatebox[origin=c]{90}{\shortstack[c]{\strut Spiral \strut}}} & Final L2 $\downarrow$ & $\num{0.325}^{+\num{0.236}}_{-\num{0.159}}$ & $\num{0.056}^{+\num{0.061}}_{-\num{0.031}}$ & {\boldmath $\mathbf{0.033}^{+\mathbf{0.016}}_{-\mathbf{0.016}}$} & $\num{0.228}^{+\num{0.041}}_{-\num{0.079}}$ & $\num{0.122}^{+\num{0.020}}_{-\num{0.031}}$ & $\num{0.063}^{+\num{0.081}}_{-\num{0.047}}$ \\
 & Cumul. L2 $\downarrow$ & $\num{28.269}^{+\num{23.817}}_{-\num{19.836}}$ & $\num{5.732}^{+\num{6.806}}_{-\num{2.495}}$ & {\boldmath $\mathbf{2.605}^{+\mathbf{1.451}}_{-\mathbf{0.480}}$} & $\num{13.834}^{+\num{3.687}}_{-\num{6.170}}$ & $\num{7.788}^{+\num{1.644}}_{-\num{2.293}}$ & $\num{6.926}^{+\num{11.042}}_{-\num{4.906}}$ \\
 & DTW $\downarrow$ & $\num{0.982}^{+\num{0.479}}_{-\num{0.273}}$ & $\num{0.580}^{+\num{0.345}}_{-\num{0.183}}$ & {\boldmath $\mathbf{0.351}^{+\mathbf{0.191}}_{-\mathbf{0.067}}$} & $\num{0.732}^{+\num{0.166}}_{-\num{0.221}}$ & $\num{0.631}^{+\num{0.111}}_{-\num{0.076}}$ & $\num{0.546}^{+\num{0.387}}_{-\num{0.263}}$ \\
\cmidrule(lr){2-8}
 & MMD$^2$ $\downarrow$ & $\num{0.020}^{+\num{0.031}}_{-\num{0.018}}$ & $\num{0.001}^{+\num{0.003}}_{-\num{0.001}}$ & {\boldmath $\mathbf{0.000}^{+\mathbf{0.000}}_{-\mathbf{0.000}}$} & $\num{0.005}^{+\num{0.003}}_{-\num{0.003}}$ & $\num{0.002}^{+\num{0.001}}_{-\num{0.001}}$ & $\num{0.003}^{+\num{0.010}}_{-\num{0.003}}$ \\
 & Energy Dist. $\downarrow$ & $\num{2.867}^{+\num{3.429}}_{-\num{1.998}}$ & $\num{0.339}^{+\num{0.616}}_{-\num{0.199}}$ & {\boldmath $\mathbf{0.082}^{+\mathbf{0.089}}_{-\mathbf{0.042}}$} & $\num{1.123}^{+\num{0.478}}_{-\num{0.488}}$ & $\num{0.440}^{+\num{0.240}}_{-\num{0.220}}$ & $(\num{4.519}^{+\num{4.034}}_{-\num{4.519}})\times10^{8}$ \\
\bottomrule
\end{tabular}
}
\end{table*}

\begin{table*}[t]
\centering
\caption{\textit{Results on the benchmarking of Sec.~\ref{sec:results-benchmark}.} (Part 2/4)}
\label{tab:eval_new_v1_test_booktabs_part2}
\sisetup{group-separator={,},group-minimum-digits=4,detect-weight=true,detect-inline-weight=math}
\footnotesize
\setlength{\tabcolsep}{4pt}
\resizebox{\textwidth}{!}{%
\begin{tabular}{clcccccc}
\toprule
Environment & Metric & \texttt{s-NLL} & \texttt{m-NLL} & \texttt{m-CRPS} & \texttt{m-arNLL} & \texttt{m-arCRPS} & \texttt{m-KDE} \\
\midrule
\multirow[c]{13}{*}{\rotatebox[origin=c]{90}{\shortstack[c]{\strut Lorenz \strut}}} & Final L2 $\downarrow$ & $\num{11.642}^{+\num{20.187}}_{-\num{7.280}}$ & $\num{11.676}^{+\num{20.290}}_{-\num{7.289}}$ & $\num{12.378}^{+\num{21.322}}_{-\num{8.270}}$ & \textit{div.} & {\boldmath $\mathbf{10.755}^{+\mathbf{16.145}}_{-\mathbf{6.936}}$} & $\num{11.388}^{+\num{15.737}}_{-\num{7.501}}$ \\
 & Cumul. L2 $\downarrow$ & $\num{866.827}^{+\num{1085.481}}_{-\num{577.921}}$ & $\num{865.478}^{+\num{1100.729}}_{-\num{592.923}}$ & $\num{888.891}^{+\num{1070.867}}_{-\num{607.088}}$ & \textit{div.} & {\boldmath $\mathbf{805.358}^{+\mathbf{1090.680}}_{-\mathbf{605.800}}$} & $\num{939.147}^{+\num{993.302}}_{-\num{664.319}}$ \\
 & DTW $\downarrow$ & $\num{101.314}^{+\num{176.614}}_{-\num{74.273}}$ & $\num{101.067}^{+\num{154.610}}_{-\num{75.645}}$ & $\num{106.475}^{+\num{163.943}}_{-\num{78.152}}$ & \textit{div.} & {\boldmath $\mathbf{91.795}^{+\mathbf{158.475}}_{-\mathbf{70.442}}$} & $\num{96.209}^{+\num{183.364}}_{-\num{76.473}}$ \\
\cmidrule(lr){2-8}
 & MMD$^2$ $\downarrow$ & $\num{0.018}^{+\num{0.034}}_{-\num{0.017}}$ & $\num{0.016}^{+\num{0.049}}_{-\num{0.016}}$ & {\boldmath $\mathbf{0.002}^{+\mathbf{0.001}}_{-\mathbf{0.002}}$} & \textit{div.} & $\num{0.032}^{+\num{0.128}}_{-\num{0.031}}$ & $\num{0.060}^{+\num{0.163}}_{-\num{0.052}}$ \\
 & Energy Dist. $\downarrow$ & $\num{32.190}^{+\num{41.168}}_{-\num{21.357}}$ & $\num{27.615}^{+\num{43.562}}_{-\num{19.143}}$ & {\boldmath $\mathbf{6.585}^{+\mathbf{10.295}}_{-\mathbf{4.436}}$} & \textit{div.} & $\num{40.672}^{+\num{91.551}}_{-\num{28.389}}$ & $\num{66.410}^{+\num{87.538}}_{-\num{36.004}}$ \\
\bottomrule
\end{tabular}
}
\end{table*}

\begin{table*}[t]
\centering
\caption{\textit{Results on the benchmarking of Sec.~\ref{sec:results-benchmark}.} (Part 3/4)}
\label{tab:eval_new_v1_test_booktabs_part3}
\sisetup{group-separator={,},group-minimum-digits=4,detect-weight=true,detect-inline-weight=math}
\footnotesize
\setlength{\tabcolsep}{4pt}
\resizebox{\textwidth}{!}{%
\begin{tabular}{clcccccc}
\toprule
Environment & Metric & \texttt{s-NLL} & \texttt{m-NLL} & \texttt{m-CRPS} & \texttt{m-arNLL} & \texttt{m-arCRPS} & \texttt{m-KDE} \\
\midrule
\multirow[c]{13}{*}{\rotatebox[origin=c]{90}{\shortstack[c]{\strut Dadras \strut}}} & Final L2 $\downarrow$ & {\boldmath $\mathbf{0.625}^{+\mathbf{1.436}}_{-\mathbf{0.504}}$} & $\num{0.971}^{+\num{1.933}}_{-\num{0.808}}$ & $\num{0.747}^{+\num{1.564}}_{-\num{0.628}}$ & \textit{div.} & $\num{1.658}^{+\num{2.602}}_{-\num{1.316}}$ & $\num{1.623}^{+\num{2.702}}_{-\num{1.391}}$ \\
 & Cumul. L2 $\downarrow$ & $\num{64.177}^{+\num{142.690}}_{-\num{48.394}}$ & $\num{86.223}^{+\num{163.618}}_{-\num{72.316}}$ & {\boldmath $\mathbf{63.810}^{+\mathbf{143.738}}_{-\mathbf{53.438}}$} & \textit{div.} & $\num{165.292}^{+\num{242.559}}_{-\num{123.316}}$ & $\num{155.975}^{+\num{262.996}}_{-\num{132.129}}$ \\
 & DTW $\downarrow$ & {\boldmath $\mathbf{5.327}^{+\mathbf{13.106}}_{-\mathbf{4.150}}$} & $\num{7.868}^{+\num{16.345}}_{-\num{6.798}}$ & $\num{6.245}^{+\num{15.722}}_{-\num{5.639}}$ & \textit{div.} & $\num{14.775}^{+\num{25.840}}_{-\num{12.719}}$ & $\num{14.076}^{+\num{27.889}}_{-\num{12.537}}$ \\
\cmidrule(lr){2-8}
 & MMD$^2$ $\downarrow$ & $\num{0.003}^{+\num{0.006}}_{-\num{0.003}}$ & $\num{0.004}^{+\num{0.006}}_{-\num{0.004}}$ & {\boldmath $\mathbf{0.001}^{+\mathbf{0.001}}_{-\mathbf{0.001}}$} & \textit{div.} & $\num{0.034}^{+\num{0.013}}_{-\num{0.032}}$ & $\num{0.021}^{+\num{0.128}}_{-\num{0.021}}$ \\
 & Energy Dist. $\downarrow$ & $\num{5.273}^{+\num{8.197}}_{-\num{3.674}}$ & $\num{4.221}^{+\num{7.851}}_{-\num{3.194}}$ & {\boldmath $\mathbf{1.905}^{+\mathbf{2.392}}_{-\mathbf{1.287}}$} & \textit{div.} & $\num{19.186}^{+\num{14.491}}_{-\num{15.053}}$ & $\num{14.222}^{+\num{40.953}}_{-\num{10.325}}$ \\
\bottomrule
\end{tabular}
}
\end{table*}

\begin{table*}[t]
\centering
\caption{\textit{Results on the benchmarking of Sec.~\ref{sec:results-benchmark}.} (Part 4/4)}
\label{tab:eval_new_v1_test_booktabs_part4}
\sisetup{group-separator={,},group-minimum-digits=4,detect-weight=true,detect-inline-weight=math}
\footnotesize
\setlength{\tabcolsep}{4pt}
\resizebox{\textwidth}{!}{%
\begin{tabular}{clcccccc}
\toprule
Environment & Metric & \texttt{s-NLL} & \texttt{m-NLL} & \texttt{m-CRPS} & \texttt{m-arNLL} & \texttt{m-arCRPS} & \texttt{m-KDE} \\
\midrule
\multirow[c]{13}{*}{\rotatebox[origin=c]{90}{\shortstack[c]{\strut Nose Hoover \strut}}} & Final L2 $\downarrow$ & $\num{0.402}^{+\num{0.207}}_{-\num{0.321}}$ & $\num{0.192}^{+\num{0.211}}_{-\num{0.158}}$ & {\boldmath $\mathbf{0.191}^{+\mathbf{0.148}}_{-\mathbf{0.144}}$} & \textit{div.} & $\num{0.193}^{+\num{0.149}}_{-\num{0.148}}$ & $\num{0.318}^{+\num{0.245}}_{-\num{0.284}}$ \\
 & Cumul. L2 $\downarrow$ & $\num{21.467}^{+\num{17.987}}_{-\num{16.129}}$ & $\num{14.180}^{+\num{17.436}}_{-\num{11.045}}$ & {\boldmath $\mathbf{13.527}^{+\mathbf{13.745}}_{-\mathbf{9.536}}$} & \textit{div.} & $\num{14.617}^{+\num{18.193}}_{-\num{8.045}}$ & $\num{21.230}^{+\num{38.851}}_{-\num{15.677}}$ \\
 & DTW $\downarrow$ & $\num{2.316}^{+\num{1.011}}_{-\num{1.992}}$ & {\boldmath $\mathbf{1.063}^{+\mathbf{1.892}}_{-\mathbf{0.877}}$} & $\num{1.172}^{+\num{0.818}}_{-\num{0.907}}$ & \textit{div.} & $\num{1.205}^{+\num{1.479}}_{-\num{0.898}}$ & $\num{1.903}^{+\num{2.600}}_{-\num{1.626}}$ \\
\cmidrule(lr){2-8}
 & MMD$^2$ $\downarrow$ & $\num{0.005}^{+\num{0.004}}_{-\num{0.005}}$ & $\num{0.002}^{+\num{0.002}}_{-\num{0.002}}$ & {\boldmath $\mathbf{0.001}^{+\mathbf{0.002}}_{-\mathbf{0.001}}$} & \textit{div.} & $\num{0.004}^{+\num{0.006}}_{-\num{0.004}}$ & $\num{0.007}^{+\num{0.016}}_{-\num{0.007}}$ \\
 & Energy Dist. $\downarrow$ & $\num{1.090}^{+\num{1.581}}_{-\num{0.890}}$ & {\boldmath $\mathbf{0.465}^{+\mathbf{0.555}}_{-\mathbf{0.383}}$} & $\num{0.488}^{+\num{0.702}}_{-\num{0.390}}$ & \textit{div.} & $\num{1.148}^{+\num{1.264}}_{-\num{0.883}}$ & $\num{1.484}^{+\num{3.402}}_{-\num{1.184}}$ \\
\bottomrule
\end{tabular}
}
\end{table*}

\clearpage

\begin{table}[t]
\centering
\caption{Training hyperparameters for the cluster sweep: model configuration.}
\label{tab:cluster_training_hparams:model}
\small
\begin{tabular}{ll}
\toprule
\textbf{Hyperparameter} & \textbf{Value} \\
\midrule
Hidden width & 64 (128 for Glycolytic oscillator) \\
Hidden layers & 3 \\
Feature dimension & 32 (64 for Glycolytic oscillator)\\
Activation & \texttt{mish} \\
Prior scale & \num{1} \\
\bottomrule
\end{tabular}
\end{table}

\begin{table}[t]
\centering
\caption{Training hyperparameters for the cluster sweep: loss-specific training settings.}
\label{tab:cluster_training_hparams:losses}
\sisetup{group-separator={,},group-minimum-digits=4}
\footnotesize
\begin{tabular}{lcccccccc}
\toprule
\textbf{Loss} & Epochs & LR & Batch & Weight decay & Grad. clip & $\lambda$  & $S$\\
\midrule
\texttt{s-NLL} & 300 & \num{0.001} & 64 $\times$ H & \num{1.0e-04} & \num{1} & $1/|\gD_{\text{single}}|$ & --  \\
\texttt{m-NLL} & 300 & \num{0.001} & 64 & \num{1.0e-04} & \num{1} & $1/|\gD_{\text{traj}}|$ & -- \\
\texttt{m-CRPS} & 300 & \num{0.001} & 64 & \num{1.0e-04} & \num{1} & $1/|\gD_{\text{traj}}|$ & 10 \\
\texttt{m-KDE} & 300 & \num{0.001} & 64 & \num{1.0e-04} & \num{1} & $1/|\gD_{\text{traj}}|$& 10 \\
\texttt{m-arNLL} & 300 & \num{0.001} & 64 & \num{1.0e-04} & \num{1}& $1/|\gD_{\text{traj}}|$ & 10 \\
\texttt{m-arCRPS} & 300 & \num{0.001} & 64 & \num{1.0e-04} & \num{1}& $1/|\gD_{\text{traj}}|$ & 10 \\
\bottomrule
\end{tabular}
\end{table}

\begin{table}[t]
\centering
\caption{Training hyperparameters for the cluster sweep: environment, data, and post-evaluation settings.}
\label{tab:cluster_training_hparams:environments}
\sisetup{group-separator={,},group-minimum-digits=4}
\scriptsize
\resizebox{\textwidth}{!}{%
\begin{tabular}{lccccccccccc}
\toprule
\textbf{Name} & Dim. & Norm. & $\Delta t$ & $\sigma$ & Traj. & Len. & Stride $s$ & $H$ & Eval steps & States & Samples $S$ for eval \\
\midrule
Spiral                       & 2 & no  & \num{0.1}   & \num{0.01} & 300 &  90 &  1 & 40 &  120 & 80 & 100 \\
Swimmer                      & 2 & no  & \num{0.1}   & \num{0.01} & 300 &  90 &  1 & 40 &  120 & 80 & 100 \\
Variable-speed limit cycle   & 2 & no  & \num{0.1}   & \num{0.01} & 300 &  90 &  1 & 40 &  120 & 80 & 100 \\
Bifurcating limit cycle      & 2 & no  & \num{0.05}  & \num{0.03} & 300 & 90 &  1 & 40 & 120 & 80 & 100 \\
Glycolytic oscillator        & 7 & yes & \num{0.01}  & \num{0.01} & 300 & 90 & 1 &  40 & 120 &  80 &  100 \\
Lorenz                       & 3 & yes & \num{0.01}  & \num{0.01} & 400 & 90 & 20 & 40 & 120 &  80 &  100 \\
Rossler                      & 3 & yes & \num{0.02}  & \num{0.01} & 400 & 90 & 20 & 40 & 120 &  80 &  100 \\
\bottomrule
\end{tabular}
}
\end{table}

\begin{table}[t]
\centering
\caption{Training hyperparameters for the cluster sweep: Wheelbot model configuration.}
\label{tab:cluster_training_hparams:wheelbot_model}
\small
\begin{tabular}{ll}
\toprule
\textbf{Hyperparameter} & \textbf{Value} \\
\midrule
Hidden width & 128 \\
Hidden layers & 3 \\
Feature dimension & 64 \\
Activation & \texttt{mish} \\
Prior scale & \num{1} \\
\bottomrule
\end{tabular}
\end{table}

\begin{table}[t]
\centering
\caption{Training hyperparameters for the cluster sweep: Wheelbot loss-specific training settings.}
\label{tab:cluster_training_hparams:wheelbot_losses}
\sisetup{group-separator={,},group-minimum-digits=4}
\footnotesize
\begin{tabular}{lccccccccc}
\toprule
\textbf{Loss} & Epochs & LR & Batch & Weight decay & Grad. clip & $H$ & Stride $s$ & $S$ & $S_{\mathrm{val}}$ \\
\midrule
\texttt{s-NLL} & 400 & \num{5.0e-04} & 256 & \num{1.0e-04} & \num{1} & -- & -- & -- & -- \\
\texttt{m-NLL} & 400 & \num{5.0e-04} & 256 & \num{1.0e-04} & \num{1} & 40 & 20 & 10 & 100 \\
\texttt{m-CRPS} & 400 & \num{5.0e-04} & 256 & \num{1.0e-04} & \num{1} & 40 & 20 & 10 & 100 \\
\texttt{m-KDE} & 400 & \num{5.0e-04} & 256 & \num{1.0e-04} & \num{1} & 40 & 20 & 10 & 100 \\
\texttt{m-arNLL} & 400 & \num{5.0e-04} & 256 & \num{1.0e-04} & \num{1} & 40 & 20 & 10 & 100 \\
\texttt{m-arCRPS} & 400 & \num{5.0e-04} & 256 & \num{1.0e-04} & \num{1} & 40 & 20 & 10 & 100 \\
\bottomrule
\end{tabular}
\end{table}

\clearpage
\section*{NeurIPS Paper Checklist}

\begin{enumerate}

\item {\bf Claims}
    \item[] Question: Do the main claims made in the abstract and introduction accurately reflect the paper's contributions and scope?
    \item[] Answer: \answerYes{} %
    \item[] Justification: All the claims from the introduction are backed by empirical evidence mainly provided in Section~\ref{sec:result} as well as in ablations and additional results in Appendix~\ref{sec:app-additional-results}.
    \item[] Guidelines:
    \begin{itemize}
        \item The answer NA means that the abstract and introduction do not include the claims made in the paper.
        \item The abstract and/or introduction should clearly state the claims made, including the contributions made in the paper and important assumptions and limitations. A No or NA answer to this question will not be perceived well by the reviewers. 
        \item The claims made should match theoretical and experimental results, and reflect how much the results can be expected to generalize to other settings. 
        \item It is fine to include aspirational goals as motivation as long as it is clear that these goals are not attained by the paper. 
    \end{itemize}

\item {\bf Limitations}
    \item[] Question: Does the paper discuss the limitations of the work performed by the authors?
    \item[] Answer: \answerYes{} %
    \item[] Justification: We explicitly include Section~\ref{sec:lims} in the main paper to thoroughly discuss the limitations of this paper and the presented approaches.
    \item[] Guidelines:
    \begin{itemize}
        \item The answer NA means that the paper has no limitation while the answer No means that the paper has limitations, but those are not discussed in the paper. 
        \item The authors are encouraged to create a separate "Limitations" section in their paper.
        \item The paper should point out any strong assumptions and how robust the results are to violations of these assumptions (e.g., independence assumptions, noiseless settings, model well-specification, asymptotic approximations only holding locally). The authors should reflect on how these assumptions might be violated in practice and what the implications would be.
        \item The authors should reflect on the scope of the claims made, e.g., if the approach was only tested on a few datasets or with a few runs. In general, empirical results often depend on implicit assumptions, which should be articulated.
        \item The authors should reflect on the factors that influence the performance of the approach. For example, a facial recognition algorithm may perform poorly when image resolution is low or images are taken in low lighting. Or a speech-to-text system might not be used reliably to provide closed captions for online lectures because it fails to handle technical jargon.
        \item The authors should discuss the computational efficiency of the proposed algorithms and how they scale with dataset size.
        \item If applicable, the authors should discuss possible limitations of their approach to address problems of privacy and fairness.
        \item While the authors might fear that complete honesty about limitations might be used by reviewers as grounds for rejection, a worse outcome might be that reviewers discover limitations that aren't acknowledged in the paper. The authors should use their best judgment and recognize that individual actions in favor of transparency play an important role in developing norms that preserve the integrity of the community. Reviewers will be specifically instructed to not penalize honesty concerning limitations.
    \end{itemize}

\item {\bf Theory assumptions and proofs}
    \item[] Question: For each theoretical result, does the paper provide the full set of assumptions and a complete (and correct) proof?
    \item[] Answer: \answerYes{} %
    \item[] Justification: For Proposition~\rev{\ref{prop:linear_exactness_rewrite}}, we include the full proof in Appendix~\ref{sec:proof_prop} and list the assumptions for the statement in the Proposition itself. Since Corollary~\ref{cor:selbo_melbo_equiv} is based on a simple sum over equivalent terms, we do not provide a full proof here.
    \item[] Guidelines:
    \begin{itemize}
        \item The answer NA means that the paper does not include theoretical results. 
        \item All the theorems, formulas, and proofs in the paper should be numbered and cross-referenced.
        \item All assumptions should be clearly stated or referenced in the statement of any theorems.
        \item The proofs can either appear in the main paper or the supplemental material, but if they appear in the supplemental material, the authors are encouraged to provide a short proof sketch to provide intuition. 
        \item Inversely, any informal proof provided in the core of the paper should be complemented by formal proofs provided in appendix or supplemental material.
        \item Theorems and Lemmas that the proof relies upon should be properly referenced. 
    \end{itemize}

    \item {\bf Experimental result reproducibility}
    \item[] Question: Does the paper fully disclose all the information needed to reproduce the main experimental results of the paper to the extent that it affects the main claims and/or conclusions of the paper (regardless of whether the code and data are provided or not)?
    \item[] Answer: \answerYes{} %
    \item[] Justification: We list the most relevant information in the main body of the text in the introduction to Section~\ref{sec:result}. Furthermore, we give more details in Appendix~\ref{sec:details} and list all the necessary hyperparameters in Tables~\ref{tab:cluster_training_hparams:model}--\ref{tab:cluster_training_hparams:wheelbot_losses}.
    \item[] Guidelines:
    \begin{itemize}
        \item The answer NA means that the paper does not include experiments.
        \item If the paper includes experiments, a No answer to this question will not be perceived well by the reviewers: Making the paper reproducible is important, regardless of whether the code and data are provided or not.
        \item If the contribution is a dataset and/or model, the authors should describe the steps taken to make their results reproducible or verifiable. 
        \item Depending on the contribution, reproducibility can be accomplished in various ways. For example, if the contribution is a novel architecture, describing the architecture fully might suffice, or if the contribution is a specific model and empirical evaluation, it may be necessary to either make it possible for others to replicate the model with the same dataset, or provide access to the model. In general. releasing code and data is often one good way to accomplish this, but reproducibility can also be provided via detailed instructions for how to replicate the results, access to a hosted model (e.g., in the case of a large language model), releasing of a model checkpoint, or other means that are appropriate to the research performed.
        \item While NeurIPS does not require releasing code, the conference does require all submissions to provide some reasonable avenue for reproducibility, which may depend on the nature of the contribution. For example
        \begin{enumerate}
            \item If the contribution is primarily a new algorithm, the paper should make it clear how to reproduce that algorithm.
            \item If the contribution is primarily a new model architecture, the paper should describe the architecture clearly and fully.
            \item If the contribution is a new model (e.g., a large language model), then there should either be a way to access this model for reproducing the results or a way to reproduce the model (e.g., with an open-source dataset or instructions for how to construct the dataset).
            \item We recognize that reproducibility may be tricky in some cases, in which case authors are welcome to describe the particular way they provide for reproducibility. In the case of closed-source models, it may be that access to the model is limited in some way (e.g., to registered users), but it should be possible for other researchers to have some path to reproducing or verifying the results.
        \end{enumerate}
    \end{itemize}

\item {\bf Open access to data and code}
    \item[] Question: Does the paper provide open access to the data and code, with sufficient instructions to faithfully reproduce the main experimental results, as described in supplemental material?
    \item[] Answer: \answerYes{} %
    \item[] Justification: All benchmarks including the real-world data are publicly available and we provide instructions to reproduce all experiments.
    The code is publicly available under the MIT license under \url{https://github.com/brunzema/gibbs-vbll}.%
    \item[] Guidelines:
    \begin{itemize}
        \item The answer NA means that paper does not include experiments requiring code.
        \item Please see the NeurIPS code and data submission guidelines (\url{https://nips.cc/public/guides/CodeSubmissionPolicy}) for more details.
        \item While we encourage the release of code and data, we understand that this might not be possible, so “No” is an acceptable answer. Papers cannot be rejected simply for not including code, unless this is central to the contribution (e.g., for a new open-source benchmark).
        \item The instructions should contain the exact command and environment needed to run to reproduce the results. See the NeurIPS code and data submission guidelines (\url{https://nips.cc/public/guides/CodeSubmissionPolicy}) for more details.
        \item The authors should provide instructions on data access and preparation, including how to access the raw data, preprocessed data, intermediate data, and generated data, etc.
        \item The authors should provide scripts to reproduce all experimental results for the new proposed method and baselines. If only a subset of experiments are reproducible, they should state which ones are omitted from the script and why.
        \item At submission time, to preserve anonymity, the authors should release anonymized versions (if applicable).
        \item Providing as much information as possible in supplemental material (appended to the paper) is recommended, but including URLs to data and code is permitted.
    \end{itemize}

\item {\bf Experimental setting/details}
    \item[] Question: Does the paper specify all the training and test details (e.g., data splits, hyperparameters, how they were chosen, type of optimizer, etc.) necessary to understand the results?
    \item[] Answer: \answerYes{} %
    \item[] Justification: As stated above, we list the most relevant information in the main body of the text in the introduction to Section~\ref{sec:result}. Furthermore, we give more details, \eg on the type of optimizer, in Appendix~\ref{sec:details} and list all the necessary hyperparameters in Tables~\ref{tab:cluster_training_hparams:model}--\ref{tab:cluster_training_hparams:wheelbot_losses}.
    \item[] Guidelines:
    \begin{itemize}
        \item The answer NA means that the paper does not include experiments.
        \item The experimental setting should be presented in the core of the paper to a level of detail that is necessary to appreciate the results and make sense of them.
        \item The full details can be provided either with the code, in appendix, or as supplemental material.
    \end{itemize}

\item {\bf Experiment statistical significance}
    \item[] Question: Does the paper report error bars suitably and correctly defined or other appropriate information about the statistical significance of the experiments?
    \item[] Answer: \answerYes{} %
    \item[] Justification: All performance plots include quantiles. Furthermore, the statistical significance of the results in Table~\ref{tab:wheelbot_training_wheelbot_all_losses_v3_booktabs} is provided by a Wilcoxon signed-rank test and the results in Figure~\ref{fig:vslc} (bottom) shows the full distribution over different initial conditions via box plots.
    \item[] Guidelines:
    \begin{itemize}
        \item The answer NA means that the paper does not include experiments.
        \item The authors should answer "Yes" if the results are accompanied by error bars, confidence intervals, or statistical significance tests, at least for the experiments that support the main claims of the paper.
        \item The factors of variability that the error bars are capturing should be clearly stated (for example, train/test split, initialization, random drawing of some parameter, or overall run with given experimental conditions).
        \item The method for calculating the error bars should be explained (closed form formula, call to a library function, bootstrap, etc.)
        \item The assumptions made should be given (e.g., Normally distributed errors).
        \item It should be clear whether the error bar is the standard deviation or the standard error of the mean.
        \item It is OK to report 1-sigma error bars, but one should state it. The authors should preferably report a 2-sigma error bar than state that they have a 96\% CI, if the hypothesis of Normality of errors is not verified.
        \item For asymmetric distributions, the authors should be careful not to show in tables or figures symmetric error bars that would yield results that are out of range (e.g. negative error rates).
        \item If error bars are reported in tables or plots, The authors should explain in the text how they were calculated and reference the corresponding figures or tables in the text.
    \end{itemize}

\item {\bf Experiments compute resources}
    \item[] Question: For each experiment, does the paper provide sufficient information on the computer resources (type of compute workers, memory, time of execution) needed to reproduce the experiments?
    \item[] Answer: \answerYes{} %
    \item[] Justification: We report runtime in Table~\ref{tab:wheelbot_training_wheelbot_all_losses_v3_booktabs} for the Mini Wheelbot and include a footnote on the type of compute used to produce the results. Still, the methods and models in this paper are rather small for todays standard and all models can also be trained on a standard laptop CPU.
    \item[] Guidelines:
    \begin{itemize}
        \item The answer NA means that the paper does not include experiments.
        \item The paper should indicate the type of compute workers CPU or GPU, internal cluster, or cloud provider, including relevant memory and storage.
        \item The paper should provide the amount of compute required for each of the individual experimental runs as well as estimate the total compute. 
        \item The paper should disclose whether the full research project required more compute than the experiments reported in the paper (e.g., preliminary or failed experiments that didn't make it into the paper). 
    \end{itemize}
    
\item {\bf Code of ethics}
    \item[] Question: Does the research conducted in the paper conform, in every respect, with the NeurIPS Code of Ethics \url{https://neurips.cc/public/EthicsGuidelines}?
    \item[] Answer: \answerYes{} %
    \item[] Justification: The paper does not violate any of the guidelines in the NeurIPS Code of Ethics.
    \item[] Guidelines:
    \begin{itemize}
        \item The answer NA means that the authors have not reviewed the NeurIPS Code of Ethics.
        \item If the authors answer No, they should explain the special circumstances that require a deviation from the Code of Ethics.
        \item The authors should make sure to preserve anonymity (e.g., if there is a special consideration due to laws or regulations in their jurisdiction).
    \end{itemize}

\item {\bf Broader impacts}
    \item[] Question: Does the paper discuss both potential positive societal impacts and negative societal impacts of the work performed?
    \item[] Answer: \answerYes{} %
    \item[] Justification:  As the paper develops general-purpose loss functions primarily in the context of dynamics learning, we believe the direct societal impact to be limited in scope, though we acknowledge relevance to safety-critical control applications. 
    \item[] Guidelines:
    \begin{itemize}
        \item The answer NA means that there is no societal impact of the work performed.
        \item If the authors answer NA or No, they should explain why their work has no societal impact or why the paper does not address societal impact.
        \item Examples of negative societal impacts include potential malicious or unintended uses (e.g., disinformation, generating fake profiles, surveillance), fairness considerations (e.g., deployment of technologies that could make decisions that unfairly impact specific groups), privacy considerations, and security considerations.
        \item The conference expects that many papers will be foundational research and not tied to particular applications, let alone deployments. However, if there is a direct path to any negative applications, the authors should point it out. For example, it is legitimate to point out that an improvement in the quality of generative models could be used to generate deepfakes for disinformation. On the other hand, it is not needed to point out that a generic algorithm for optimizing neural networks could enable people to train models that generate Deepfakes faster.
        \item The authors should consider possible harms that could arise when the technology is being used as intended and functioning correctly, harms that could arise when the technology is being used as intended but gives incorrect results, and harms following from (intentional or unintentional) misuse of the technology.
        \item If there are negative societal impacts, the authors could also discuss possible mitigation strategies (e.g., gated release of models, providing defenses in addition to attacks, mechanisms for monitoring misuse, mechanisms to monitor how a system learns from feedback over time, improving the efficiency and accessibility of ML).
    \end{itemize}
    
\item {\bf Safeguards}
    \item[] Question: Does the paper describe safeguards that have been put in place for responsible release of data or models that have a high risk for misuse (e.g., pretrained language models, image generators, or scraped datasets)?
    \item[] Answer: \answerNA{} %
    \item[] Justification: The paper proposed general loss functions for regression. As such, the paper poses no such risks.
    \item[] Guidelines:
    \begin{itemize}
        \item The answer NA means that the paper poses no such risks.
        \item Released models that have a high risk for misuse or dual-use should be released with necessary safeguards to allow for controlled use of the model, for example by requiring that users adhere to usage guidelines or restrictions to access the model or implementing safety filters. 
        \item Datasets that have been scraped from the Internet could pose safety risks. The authors should describe how they avoided releasing unsafe images.
        \item We recognize that providing effective safeguards is challenging, and many papers do not require this, but we encourage authors to take this into account and make a best faith effort.
    \end{itemize}

\item {\bf Licenses for existing assets}
    \item[] Question: Are the creators or original owners of assets (e.g., code, data, models), used in the paper, properly credited and are the license and terms of use explicitly mentioned and properly respected?
    \item[] Answer: \answerYes{} %
    \item[] Justification: All models were build in Flax NNX which is available under the Apache-2.0 license. Furthermore, we use the chaotic systems from the \texttt{dysts} library~\citep{gilpin2chaos} which is also available under the Apache-2.0 license. 
    We calculate the DTW using \texttt{dtaidistance} available under the Apache-2.0 license.
    The Mini Wheelbot dataset is available under the MIT license.
    \item[] Guidelines:
    \begin{itemize}
        \item The answer NA means that the paper does not use existing assets.
        \item The authors should cite the original paper that produced the code package or dataset.
        \item The authors should state which version of the asset is used and, if possible, include a URL.
        \item The name of the license (e.g., CC-BY 4.0) should be included for each asset.
        \item For scraped data from a particular source (e.g., website), the copyright and terms of service of that source should be provided.
        \item If assets are released, the license, copyright information, and terms of use in the package should be provided. For popular datasets, \url{paperswithcode.com/datasets} has curated licenses for some datasets. Their licensing guide can help determine the license of a dataset.
        \item For existing datasets that are re-packaged, both the original license and the license of the derived asset (if it has changed) should be provided.
        \item If this information is not available online, the authors are encouraged to reach out to the asset's creators.
    \end{itemize}

\item {\bf New assets}
    \item[] Question: Are new assets introduced in the paper well documented and is the documentation provided alongside the assets?
    \item[] Answer: \answerNA{} %
    \item[] Justification: The paper does not provide additional assets such as new datasets but builds on existing publicly available simulations and real-world data.
    \item[] Guidelines:
    \begin{itemize}
        \item The answer NA means that the paper does not release new assets.
        \item Researchers should communicate the details of the dataset/code/model as part of their submissions via structured templates. This includes details about training, license, limitations, etc. 
        \item The paper should discuss whether and how consent was obtained from people whose asset is used.
        \item At submission time, remember to anonymize your assets (if applicable). You can either create an anonymized URL or include an anonymized zip file.
    \end{itemize}

\item {\bf Crowdsourcing and research with human subjects}
    \item[] Question: For crowdsourcing experiments and research with human subjects, does the paper include the full text of instructions given to participants and screenshots, if applicable, as well as details about compensation (if any)? 
    \item[] Answer: \answerNA{} %
    \item[] Justification: Our paper does not involve crowdsourcing nor research with human subjects.
    \item[] Guidelines:
    \begin{itemize}
        \item The answer NA means that the paper does not involve crowdsourcing nor research with human subjects.
        \item Including this information in the supplemental material is fine, but if the main contribution of the paper involves human subjects, then as much detail as possible should be included in the main paper. 
        \item According to the NeurIPS Code of Ethics, workers involved in data collection, curation, or other labor should be paid at least the minimum wage in the country of the data collector. 
    \end{itemize}

\item {\bf Institutional review board (IRB) approvals or equivalent for research with human subjects}
    \item[] Question: Does the paper describe potential risks incurred by study participants, whether such risks were disclosed to the subjects, and whether Institutional Review Board (IRB) approvals (or an equivalent approval/review based on the requirements of your country or institution) were obtained?
    \item[] Answer: \answerNA{} %
    \item[] Justification: Our paper does not involve crowdsourcing nor research with human subjects.
    \item[] Guidelines:
    \begin{itemize}
        \item The answer NA means that the paper does not involve crowdsourcing nor research with human subjects.
        \item Depending on the country in which research is conducted, IRB approval (or equivalent) may be required for any human subjects research. If you obtained IRB approval, you should clearly state this in the paper. 
        \item We recognize that the procedures for this may vary significantly between institutions and locations, and we expect authors to adhere to the NeurIPS Code of Ethics and the guidelines for their institution. 
        \item For initial submissions, do not include any information that would break anonymity (if applicable), such as the institution conducting the review.
    \end{itemize}

\item {\bf Declaration of LLM usage}
    \item[] Question: Does the paper describe the usage of LLMs if it is an important, original, or non-standard component of the core methods in this research? Note that if the LLM is used only for writing, editing, or formatting purposes and does not impact the core methodology, scientific rigorousness, or originality of the research, declaration is not required.
    \item[] Answer: \answerNA{} %
    \item[] Justification: The paper does not include LLMs in any part of the method.
    \item[] Guidelines:
    \begin{itemize}
        \item The answer NA means that the core method development in this research does not involve LLMs as any important, original, or non-standard components.
        \item Please refer to our LLM policy (\url{https://neurips.cc/Conferences/2025/LLM}) for what should or should not be described.
    \end{itemize}

\end{enumerate}

\end{document}